\documentclass[a4paper]{article}
\usepackage[it]{idsiatechrep}   % italian acronyms

\usepackage{times}
\usepackage{graphicx}
\usepackage{color}
\usepackage{url}
\usepackage[authoryear]{natbib}

\usepackage[cmex10]{amsmath}

\usepackage{amsthm}  %proofs and lemmas
\newcommand{\avt}[1]{\langle #1 \rangle}

\renewcommand{\thesubsubsection}{\arabic{section}.\arabic{subsubsection}}

\newcommand{\captionfonts}{\normalsize}

\makeatletter \long\def\@makecaption#1#2{%
  \vskip\abovecaptionskip
  \sbox\@tempboxa{{\captionfonts #1: #2}}%
  \ifdim \wd\@tempboxa >\hsize
    {\captionfonts #1: #2\par}
  \else
    \hbox to\hsize{\hfil\box\@tempboxa\hfil}%
  \fi
  \vskip\belowcaptionskip}
\makeatother %%%%%

\title{Incremental Slow Feature Analysis: Adaptive and Episodic Learning from High-Dimensional Input Streams}
\author{Varun R. Kompella, Matthew Luciw and J$\ddot{\mbox{u}}$ergen Schmidhuber }
\date{November 2011}

\reportnumber{07-11}

\begin{document}
\makecover         % makes the cover sheet
\maketitle
%Abstract
\begin{center} {\bf Abstract} \end{center}
Slow Feature Analysis (SFA) extracts features representing the underlying causes of changes within a temporally coherent high-dimensional raw sensory input signal.  Our novel incremental version of SFA (IncSFA) combines incremental Principal Components Analysis and Minor Components Analysis.  Unlike standard batch-based SFA, IncSFA adapts along with non-stationary environments, is amenable to episodic training, is not corrupted by outliers, and is covariance-free.  These properties make IncSFA a generally useful unsupervised preprocessor for autonomous learning agents and robots.  In IncSFA, the CCIPCA and MCA updates take the form of Hebbian and anti-Hebbian updating, extending the biological plausibility of SFA.   In both single node and deep network versions, IncSFA learns to encode its input streams (such as high-dimensional video) by informative slow features representing meaningful abstract environmental properties. It can handle cases where batch SFA fails.

\section{Introduction}

%what sfa is... where it comes from... what its been used for
Slow feature analysis~\citep{WisSej2002,sfascholar}(SFA) is an unsupervised learning technique that extracts features from an input stream with the objective of maintaining an informative but slowly-changing feature response over time.  The idea of using~\textit{temporal stability} as an objective in learning systems has motivated some other unsupervised learning techniques~\citep{Hinton1989,földiák1991learning,mitchison1991removing,Schmidhuber:92ncchunker,NIPS2009_0933}.   SFA is distinguished by its formulation of the feature extraction problem as an eigensystem problem, which guarantees that its solution methods reliably converge to the best solution, given its constraints (no local minima problem).  SFA has shown success in problems such as extraction of driving forces of a dynamical system~\citep{wiskott2003estimating}, nonlinear blind source separation~\citep{sprekeler2010extension}, as a preprocessor for reinforcement learning~\citep{legenstein2010reinforcement,AutoIncSFA2011}, and learning of place-cells, head-direction cells, grid-cells, and spatial view cells from high-dimensional visual input~\citep{franzius2007slowness} --- such representations also exist in biological agents~\citep{o1971hippocampus,taube1990head,rolls1999spatial,hafting2005microstructure}.

%Despite SFA's promise, it has not reached widespread application.

There are limitations to existing SFA implementations due to their batch processing nature, which becomes especially apparent when attempting to apply it in somewhat uncontrolled environments.  To overcome these issues, we introduce the new Incremental Slow Feature Analysis (IncSFA)~\citep{kompellaincremental,AutoIncSFA2011}.  A few earlier techniques with temporal continuity objective  were incremental as well~\citep{Hinton1989,NIPS2009_0933}, but IncSFA follows the SFA formulation and can track solutions of batch SFA (BSFA), over which it has the following advantages:

\begin{itemize}
\item \textbf{Adaptation to changing input statistics.}  BSFA requires all data to be collected in advance. New data cannot be used to modify already learned slow features. Once the input statistics change, IncSFA can automatically adapt its features without outside intervention, while BSFA has to discard previous features to process the new data.

    In open-ended learning settings, an autonomous agent's lifelong input stream follows such a nonstationary distribution. The agent's behavior will typically change over time, thus generating new input sequences.  Features useful for early behaviors may not be useful for later behavior.

    %so, if we want the agent to autonomously develop, it is necessary for the features themselves to have the capacity to adapt along with the agent's behavior.
\item \textbf{Learn features across episodes.}  Episodic learning is impossible for BSFA, since it cannot handle temporal discontinuities at episode boundaries.  IncSFA, however,  may use the final slow features from the previous episode
   to  initialize its features of the next episode.
\item \textbf{Reduced sensitivity to outliers.}   Real-world environments typically exhibit infrequent, uncontrolled, insignificant external events that should be ignored. BSFA is very sensitive to  such events, encoding everything that changes slowly within the current  batch.
%juergen: below you already mention results here without even defining what's a first slow feature... this whole item is confusing and disturbing and too long - the plasticity argument is unclear here - if one cannot express compactly what's the problem, then probably it's not worth it...
%For example, in a set of images with a moving interactor agent in the foreground, a door in the background is opened accidently over just a few frames. The first slow feature found by BSFA was found to code for this door opening event.
IncSFA's plasticity, however, makes it lose sensitivity to such events over time.
\item \textbf{Covariance-free.} BSFA techniques rely upon batch Principal Component Analysis~\citep{Jolliffe} (PCA), which requires the data's covariance matrix.  Estimating, storing and/or updating covariance matrices can be expensive for high-dimensional data and impractical for open-ended learning.  IncSFA uses  \textit{covariance-free} techniques.   For high-dimensional images, the number of parameters to estimate in the covariance matrix is huge: $n(n+1)/2$ for dimension $n$, while a covariance-free technique only requires $n \times m$ where $m$ is the desired number of principal components (PCs).  For example, $100 \times 100$ dimensional images lead to $50,005,000$ free parameters in the covariance matrix,  which is reduced to only $100 \times 100 \times m$ eigenvector parameters with covariance-free updating.

    Furthermore, since often only a relatively small number of principal components are needed to explain most of the variance in the data, the other components do not even have to be estimated.  With IncSFA, dimensionality reduction can be done during  PC estimation; no time needs to be wasted on computing the many insignificant lower-order PCs.

    Another technical problem with the covariance matrix: in sequences where only a small part of the input  changes, computing the principal components of the difference signal's covariance matrix will result in singularity errors, since the matrix won't have full rank.
\item \textbf{Biological Plausibility.}  IncSFA adds further biological plausibility to SFA.  SFA itself is linkable to biological systems due to the results in deriving place cell, grid cells, etc., but it is difficult to see how BSFA could be realized in the brain.  IncSFA's updates, however, can be described in incremental Hebbian and anti-Hebbian terms.
\end{itemize}

The remainder of this paper is organized as follows.  Section~\ref{SE:BG} reviews SFA and its batch solution.  Section~\ref{SE:INCSFA} describes the new incremental SFA.  Section~\ref{SE:ALGO} details the algorithm and discusses deeper related issues, including convergence conditions and parameter setting. Section~\ref{SE:RESULTS} contains experiments and results, and shows how to utilize IncSFA as part of a hierarchical image processing architecture.  Section~\ref{SE:CONCLUDE} concludes the paper.

\section{Background}
\label{SE:BG}

\subsection{SFA: Intuition}

\begin{figure}[!ht]
%\begin{center}
%\hspace{-0.55in}
\includegraphics[width=15cm]{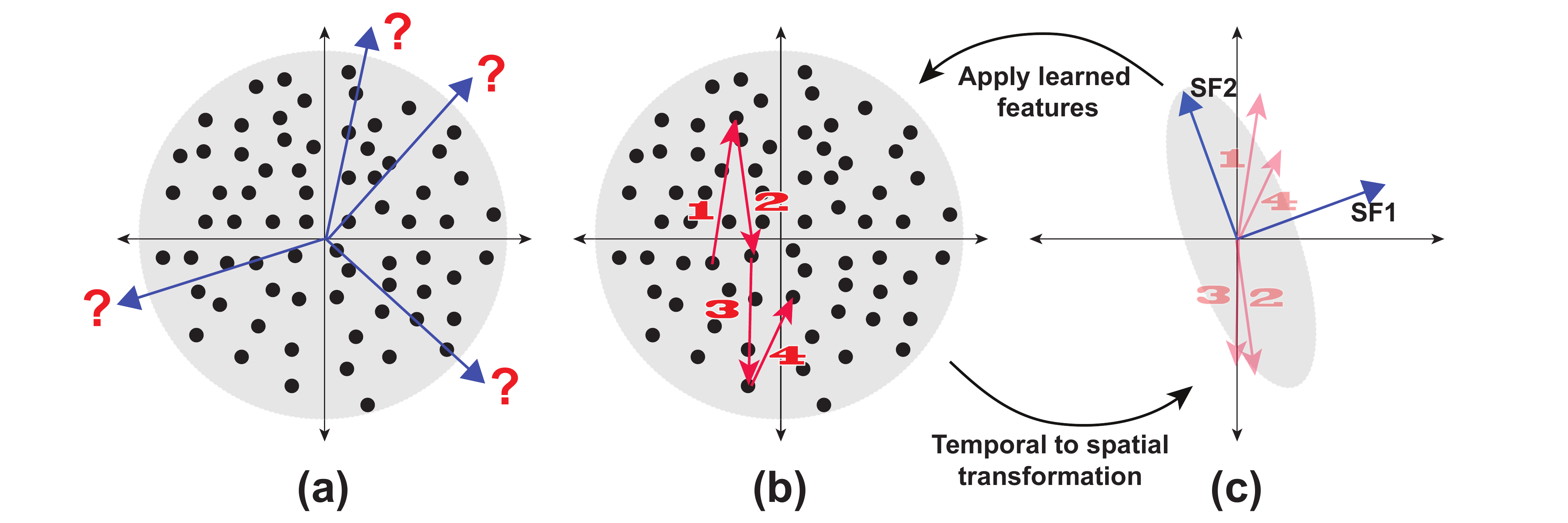}
%\end{center}
\caption{Intuition of SFA.  (A): Consider a zero-mean input signal that spatially resembles white noise.  Assume the input distributions are Gaussian, shown by the gray area, while the black dots show individual data points.  Spatial feature extractors such as PCA will not prefer any direction over any other.   (B):  Eschewing unhelpful spatial processing, we examine this input as a time-series; to illustrate here we show a short sequence of input.  Each difference vector becomes a spatial component in the space shown in (C).  In this space, the first principal component gives the (linear) direction of quickest change.  The second --- the minor component --- gives the direction of slowest change.  We see that recoding the data in terms of subsequent differences and performing an eigendecomposition provides an ordered set of separate features, which are applied to the original input signal.}
\label{fig:sfaexample}
\end{figure}

%what SFA is (intuitively)

We first review SFA briefly in an intuitive sense.  SFA is a form of unsupervised learning (UL). It searches  for a set of mappings $g_i$from data ${\bf x} \in \mathcal{R}^I$ to output components $y_i = g_i({\bf x})$ that are \textit{separate} from each other in some sense and express information that is in some sense \textit{relevant}.  In SFA separateness is realized as decorrelation (like in PCA), while relevance is defined in terms of slowness of change over time.  Ordering our functions $g_1,g_2,...,g_I$ by slowness, we can discard all but the $J < I$ slowest, to enable dimensionality reduction, getting rid of irrelevant information such as quickly changing noise  assumed to be useless.  The compact relevant data encodings reduce the search space for downstream goal-directed learning procedures~\citep{Schmidhuber:99zif,barlow2001redundancy}. As an example, consider a high-dimensional dynamical system: a mobile robot sensing with an onboard camera, where each pixel is considered a separate observation component.  SFA will use the video sequences to guide its search over functions that encode each image into a small set of state variables, and the robot can use these new state variables to quickly develop useful controllers. Fig.~\ref{fig:sfaexample} provides a visual example of how SFA operates.

SFA-based UL learns \textit{instantaneous} features from sequential data~\citep{Hinton1989,WisSej2002,doerschtemporal}.   Relevance cannot be \textit{uncovered} without taking time into account, but once it is known, each input frame can be processed on its own.  SFA differs  from both 1. many well-known unsupervised feature extractors~\citep{Abut90,Jolliffe,Comon94,lee1999learning,Kohonen01,hinton:2002}, which ignore dynamics, and 2. Other UL systems that both learn and apply features to sequences~\citep{Schmidhuber:92ncchunker,Schmidhuber:92nips,Schmidhuber:92ncfactorial,Steffi:93cmss,Klapper:01,jenkins2004spatio,lee2010unsupervised,BigDog2011agi}, thus assuming that the state of the system itself can depend on past information.

%A slowly changing feature output is temporally stable.  There is a link between temporal stability and predictive coding.  It has been shown, under some assumptions, that SFA seeks to maximize the mutual information between the current feature response and the next input~\citep{shaw2005predictive,creutzig2008predictive}.

\subsection{SFA: Formulation}

SFA's optimization problem~\citep{WisSej2002,franzius2007slowness} is formally written as follows:

\textit{Given an I-dimensional sequential input signal ${\bf x}(t) = [x_{1}(t), ..., x_{I}(t)]^{T}$, find a set of $J$ instantaneous real-valued functions ${\bf g}(x) = [g_{1}({\bf x}), ...,g_{J}({\bf x})]^{T}$, which
together generate a $J$-dimensional output signal ${\bf y}(t) = [y_{1}(t), ...,y_{J}(t)]^{T}$ with $y_{j}(t) := g_{j}({\bf x}(t))$, such
that for each $j \in \{1, ...,J\}$
\begin{equation}
\Delta_{j} := \Delta(y_{j}):=\avt{\dot{y}_{j}^{2}} \quad\textrm{is minimal} \label{eqn:Delta}
\end{equation}
under the constraints
\begin{eqnarray}
\avt{y_{j}} & = & 0\quad\textrm{(zero mean),}\label{eqn:zero mean old}\\
\avt{y_{j}^{2}} & = & 1\quad\textrm{(unit variance),}\label{eqn:unit variance old}\\
\forall i<j:\avt{y_{i}y_{j}} & = & 0\quad\textrm{(decorrelation and order),}\label{eqn:decorrelation old}
\end{eqnarray}
with $\avt{\cdot}$ and $\dot{y}$ indicating temporal averaging and the derivative of $y$, respectively.}

The problem is to find instantaneous functions $g_j$ that generate different output signals varying as slowly as possible.    The constraints (\ref{eqn:zero mean old}) and (\ref{eqn:unit variance old}) together avoid a trivial constant output solution.  The decorrelation constraint (\ref{eqn:decorrelation old}) ensures that different functions $g_j$ do not code for the same features.

%Decorrelation is weaker than independence, but independence is more difficult to realize.

\subsection{Batch SFA}

Solving this learning problem involves non-trivial variational calculus optimization. But it is simplified through an eigenvector approach.  If the $g_j$ are linear combinations of a finite set of nonlinear functions ${\bf h}$, then

\begin{equation}
\label{EQ:SFA1}
y_j(t) = g_j({\bf x}(t)) = {\bf w}^T_j ~ {\bf h}({\bf x}(t)) = {\bf w}^T_j ~ {\bf z}(t),
\end{equation}

%\noindent and let ${\bf z}(t) = {\bf h}({\bf x}(t))$.

\noindent and the SFA problem now becomes to find weight vectors ${\bf w}_j$ to minimize the rate of change of the output variables,

\begin{equation}
\label{EQ:OBJECTIVE}
\Delta(y_j) = \avt{ \dot{y}_j^2 } = {\bf w}_j^T ~ \avt{ \dot{ {\bf z}} \dot{ {\bf z}}^T } ~ {\bf w}_j,
\end{equation}

\noindent subject to the constraints (2-4).  The slow feature learning problem has become linear on the derivative signal $\dot{{\bf z}}$.

If the functions of ${\bf h}$ are chosen such that ${\bf z}$ has unit covariance matrix and zero mean, the three constraints will be fulfilled if and only if the weight vectors ${\bf w}_j$ are orthonormal.  Eq.~\ref{EQ:OBJECTIVE} will be minimized, and the orthonormal constraint satisfied, with the set of $J$ normed eigenvectors of $\avt{ \dot{ {\bf z}} \dot{ {\bf z}}^T }$ with the $J$ smallest eigenvalues (for any $J \le I$).

The BSFA technique practically implements this solution by using batch principal component analysis (PCA) \citep{Jolliffe} twice.  Referring back to Eq.~\ref{EQ:OBJECTIVE}, to select ${\bf h}$ appropriately, a well-known process called whitening (or sphering), is used to map ${\bf x}$ to a ${\bf z}$ with zero mean and identity covariance matrix, thus decorrelating signal components and scaling them so that there is unit variance along each PC direction.  Whitening serves as a bandwidth normalization, so that slowness can truly be measured (slower change will not simply be due to a low variance direction).  Whitening requires the PCs of the input signal (PCA \#1).  The orthonormal basis that minimizes the rate of output change are the minor components -- principal components with smallest eigenvalues -- in the derivative space.  So another PCA (\#2) on $ \dot{ {\bf z}}$ yields the slow features (eigenvectors) and their order (via eigenvalues).

\section{Incremental SFA}
\label{SE:INCSFA}

IncSFA also employs the eigenvector tactic, but may update an existing estimate on any amount of new data, even a single data point ${\bf x}(t)$.  A high-level formulation is

\begin{equation}
\label{EQ:INCSFAHL}
({\bf W}(t+1), \theta(t+1)) = IncSFA({\bf W}(t), {\bf x}(t), \theta(t)),
\end{equation}

\noindent where ${\bf W} = \left({\bf w}_1, ...,{\bf w}_J\right)$ is the matrix of existing slow feature vector estimates, and $\theta$ contains algorithm memory and parameters, which we will discuss later.

To replace PCA $\#1$, IncSFA needs to do online whitening of input ${\bf x}$.  We use Candid Covariance-Free Incremental (CCI) PCA~\citep{WengCCIPCA03}.   CCIPCA  incrementally updates both the eigenvectors and eigenvalues necessary for whitening, and does not keep an estimate of the covariance matrix.  CCIPCA is also used to reduce dimensionality.

Except for low-dimensional derivative signals $\dot{{\bf z}}$, CCIPCA cannot replace PCA $\#2$.  It will be unstable, since the slow features correspond to the least significant components.  Minor Components Analysis (MCA)~\citep{oja1992principal} incrementally extracts the principal components with the smallest eigenvalues.  We use Peng's low complexity updating rule~\citep{peng2007convergence}.  Peng proved its convergence even for constant learning rates---good for open-ended learning.  MCA with sequential addition~\citep{chen2001sequential,peng2006new} will extract multiple slow features in parallel.

\subsection{Neural Updating for PC and MC Extraction}
\label{SE:HISTO}

CCIPCA and Peng's MCA are the most appropriate incremental PCA and MCA algorithms for IncSFA.  To justify these choices, we briefly review the literature on neural networks that perform incremental PCA and MCA.

Well-known incremental PCA algorithms are Oja and Karhunen's Stochastic Gradient Ascent (SGA)~\citep{Oja85}, Sanger's Generalized Hebbian Algorithm (GHA)~\citep{sanger1989optimal}, and CCIPCA.  They all build on the work of Amari (1977) and Oja (1982), who showed that a linear neural unit using Hebbian updating could compute the first principal component of a data set~\citep{amari1977neural,oja1982simplified}\footnote{Much earlier work of a non-neural network flavor had shown how the first PC, including the eigenvalue could be learned incrementally~\citep{krasulina1970method}.}.  However, SGA (1985) builds upon Oja's earlier work, GHA (1989) builds upon SGA, and CCIPCA (2003) builds upon GHA.

SGA use Gram-Schmidt Orthonormalization (GSO) to incrementally find the subspace of all principal components, but there is no guarantee of finding the components themselves.  Sanger used Kreyszig's \citep{Kreyszig88} (1988) (faster/more effective) residual vector method for computing multiple components.  His provably converging GHA used the residual method for simultaneous computation of all components.  CCIPCA~\citep{WengCCIPCA03} modified GHA to be ``candid'', meaning it maintained an implicit learning rate dependant on the data, greatly increasing the algorithm's efficiency so that it became useful for high-dimensional inputs, such as in appearance-based computer vision.  This incremental PCA updating method is the best of the above for IncSFA.  It converges~\citep{zhang2001convergence} to both eigenvectors and eigenvalues, necessary since whitening requires both.  Due to its candidness, potentially difficult learning rate ``hand-tuning'' is minimized.

%While the importance of the PCs was well-known, applications where MCs were useful were traditionally few and far between, although now more have being realized.  In traditional signal processing, the minor subspace corresponded to the noise subspace. But Thomson~\citep{thompson1979adaptive} showed how MCs could enable frequency estimation of signals buried in white noise, and Xu \textit{et al.} showed its application to curve/surface fitting.

As for MCA: Xu \textit{et al.}~\citep{xu1992modified} were the first to show that a linear neural unit equipped with anti-Hebbian learning could extract minor components.  Oja modified SGA's updating method to an anti-Hebbian variant~\citep{oja1992principal}, and showed how it could converge to the MC subspace.  Studying the nature of the duality between PC and MC subspaces~\citep{wang1996unified,chen1998unified},  Chen, Amari and Lin~\citep{chen2001sequential} (2001) introduced the sequential addition technique, enabling linear networks to efficiently extract multiple MCs simultaneously.  Building upon previous MCA algorithms, Peng (2007)~\citep{peng2007convergence} derived the conditions and a learning rule for extracting MCs without changing the learning rate.  Sequential addition was added to this rule so that multiple MCs could be extracted~\citep{peng2006new}.  We use this MCA updating method since it gives us the actual minor components, not just the subspace they span, and it allows for a constant learning rate, which can be quite high, leading to a quick reasonable estimate of the true components.

\subsection{CCIPCA Updating}
\label{SE:CCIPCA}

Given zero-mean data ${\bf u} = {\bf x} - \mbox{E}[{\bf x}]$, a PC is a normed eigenvector ${\bf v}^*_i$ of the data covariance matrix $\mbox{E}[{\bf u}{\bf u}^T]$. Eigenvalue $\lambda^*_i$ is the variance of the samples along ${\bf v}^*_i$.  By definition, an eigenvector and eigenvalue satisfy

\begin{equation}
\label{EQ:EIGEN}
\mbox{E}[{\bf u}{\bf u}^T] {\bf v}^*_i = \lambda^*_i {\bf v}^*_i,
\end{equation}

The set of eigenvectors are orthonormal, and ordered such that $\lambda^*_1 \geq \lambda^*_2 \geq ... \geq \lambda^*_K$.

The whitening matrix is generated by multiplying the matrix of principal components $\hat{{\bf V}} = \left[{\bf v}^*_1,..{\bf v}^*_K \right]$ by the diagonal matrix
$\hat{{\bf D}}$, where component $\hat{d}_{i,i} = \displaystyle \frac{1}{\sqrt{\lambda^*_i}}$.  After whitening via ${\bf z}(t) = \hat{{\bf V}} \hat{{\bf D}} {\bf u}(t)$, then $\mbox{E}[{\bf z}{\bf z}^T] = I$.  In IncSFA, we use online estimates of $\hat{{\bf V}}$ and $\hat{{\bf D}}$.  Both eigenvectors and eigenvalues need to be estimated.

%\subsection{CCIPCA Updating}

CCIPCA updates ${\bf V}$ and ${\bf D}$ from each sample.  For inputs ${\bf u}_i$, the first PC is the expectation of the normalized response-weighted inputs.  Eq~\ref{EQ:EIGEN} can be rewritten as

\begin{equation}
\label{EQ:RWI}
\lambda^*_i ~ {\bf v}^*_i = \mbox{E}\left[ ({\bf u}_i \cdot {\bf v}^*_i) ~ {\bf u}_i \right],
\end{equation}

The corresponding incremental updating equation, where $\lambda^*_i ~ {\bf v}^*_i$ is estimated by ${\bf v}_i(t)$, is

\begin{equation}
\label{EQ:HEBBIANUPDATING}
{\bf v}_i(t) = \displaystyle (1 - \eta) ~ {\bf v}_i(t-1) + \eta ~ \left[ \frac{{\bf u}_i(t) \cdot {\bf v}_i(t-1) }{ \| {\bf v}_i(t-1) \| } ~ {\bf u}_i(t) \right].
\end{equation}

\noindent where $\eta$ is the learning rate.  In other words, both the eigenvector and eigenvalue of the first PC of ${\bf u}_i$ can be found through the sample mean-type updating in Eq.~\ref{EQ:RWI}.  The estimate of the eigenvalue is given by $\lambda_i = \|{\bf v}_i(t)\|$.  Using both a learning rate $\eta$ and retention rate $(1-\eta)$ automatically controls the adaptation of the vector with respect to the magnitude of the data vectors, leading to efficiency and stability.

\subsection{Lower-Order Principal Components}

Any component $i>1$ not only must satisfy Eq.~\ref{EQ:EIGEN} but must also be constrained to be orthogonal to the higher-order components.  The residual method generates observations in a complementary space so that lower-order eigenvectors can be found by the same update rule Eq.~\ref{EQ:HEBBIANUPDATING}.

Denote ${\bf u}_i(t)$ as the observation for component $i$.  When $i=1$, ${\bf u}_1(t) = {\bf u}(t)$.  When $i>1$, ${\bf u}_i$ is a
residual vector, which has the ``energy'' of ${\bf u}(t)$ from the higher-order components removed.  Solving for the first PC in this residual space solves for the $i$-th component overall.  To create a residual vector, ${\bf u}_i$ is projected onto ${\bf v}_i$ to get the energy of ${\bf u}_i$ that ${\bf v}_i$ is responsible for.  Then, the energy-weighted ${\bf v}_i$ is subtracted from ${\bf u}_i$ to obtain ${\bf u}_{i+1}$:

\begin{equation}
\label{EQ:Residual} {\bf u}_{i+1}(t) = {\bf u}_i(t) - \left( {\bf u}^T_i(t) \displaystyle
\frac{{\bf v}_i(t)}{\|{\bf v}_i(t)\|} \right) \displaystyle \frac{{\bf v}_i(t)}{\|{\bf v}_i(t)\|}.
\end{equation}

Together, Eq.~\ref{EQ:HEBBIANUPDATING} and Eq.~\ref{EQ:Residual} constitute the CCIPCA technique, which was proven to converge to the true components~\citep{zhang2001convergence}.  Yet, due to the residual method, the speed of learning is in line with the order: the first PC must be ``sufficiently correct'' before the second PC can start to learn, and so on.

\subsection{MCA Updating}
\label{SE:MCA}

After using CCIPCA components to generate an approximately whitened signal ${\bf z}$, the derivative is approximated by $\dot{{\bf z}}(t) = {\bf z}(t) - {\bf z}(t-1)$.  In this derivative space, the minor components on $\dot{{\bf z}}$ are the slow features.

To find the minor component, Peng's MCA update rule~\citep{peng2007convergence} is used,

%combines anti-Hebbian learning with a penalty term designed such that the first MC can be proved to be the only stable solution of the updating method~\citep{peng2007convergence}, without requiring learning rate to go to zero.  The incremental update rule is as follows:

\begin{eqnarray}
\label{EQ:MCAUPDATE}
{\bf w}_i(t) &=& 1.5 {\bf w}_i(t-1) - \eta ~ {\bf C}_i ~ {\bf w}_i(t-1) \\ \nonumber
             &-& \eta ~ [{\bf w}_i^T(t-1) {\bf w}_i(t-1)] ~ {\bf w}_i(t-1),
\end{eqnarray}

\noindent where, for the first minor component, ${\bf C}_1 = \dot{{\bf z}}(t) \dot{{\bf z}}^T(t)$.

For stability and convergence, the following constraints must be satisfied,

\begin{eqnarray}
\label{EQ:CONVERGECOND}
\eta \lambda_1 < 0.5, \\
||{\bf w}(0)||^2 \leq \frac{1}{2 \eta}, \\
{\bf w}^T(0) {\bf w}^* \neq 0
\end{eqnarray}

\noindent where ${\bf w}(0)$ is the initial feature estimate and ${\bf w}^*$ the true eigenvector associated with the smallest eigenvalue.  Basically, the learning rate must not be too large, and the initial estimate must not be orthogonal to the true component.

\subsection{Lower-Order Slow Features}

%Chen \textit{et al.} \citep{chen2001sequential} pointed out an interesting duality between minor components and PCs.  For some positive definite matrix ${\bf C}$, the eigenvectors of $\gamma I - {\bf C}$, where $\gamma > \lambda_1$, will be of the opposite order.  Purely switching the signs in CCIPCA will not work for extracting multiple MCs.

Sequential addition shifts each observation into a space where the minor component of the current space will be the first PC, and all other PCs are reduced in order by one.  It does this by adding the scale of the first PC to the already estimated slow feature directions.  This allows IncSFA to extract more than one slow feature in parallel.  Sequential addition updates the matrix ${\bf C}_i, ~ \forall i > 1$ as follows:

\begin{equation}
\label{EQ:SADD}
{\bf C}_i(t) = {\bf C}_{i-1}(t) + \gamma(t) ~ \left( {\bf w}_{i-1}(t) {\bf w}_{i-1}^T(t)\right) / \left( {\bf w}_{i-1}^T(t) {\bf w}_{i-1}(t)\right)
\end{equation}

%old version
%\begin{equation}
%\label{EQ:SADD}
%{\bf C}_i(t) = {\bf C}_{i-1}(t) + \gamma(t) ~ \sum^{i-1}_{j=1} {\bf w}_j(t) %{\bf w}_j^T(t) {\bf C}_1(t)
%\end{equation}

Note Eq.~\ref{EQ:SADD} introduces parameter $\gamma$, which must be larger than the largest eigenvalue of $\mbox{E}[\dot{{\bf z}}(t) \dot{{\bf z}}^T(t)]$.  To automatically set $\gamma$, we compute the greatest eigenvalue of the derivative signal through another CCIPCA rule to update only the first PC. Then, let $\gamma = \lambda_1(t) + \epsilon$ for small $\epsilon$.

\subsection{Link to Hebbian and Anti-Hebbian Updating}
\label{SE:BIO}

BSFA has been shown to derive slow features that operate like biological grid cells\footnote{A grid cell has high firing rate when the animal is in certain positions in its closed environment --- viewed from above, the pattern resembles a grid.} from quasi-natural image streams, which are recorded from the camera of a moving agent exploring an enclosure~\citep{franzius2007slowness}.  In rats, grid cells are found in entorhinal cortex (EC)~\citep{hafting2005microstructure}, which feeds into the hippocampus. Augmenting the BSFA network with an additional competitive learning (CL) layer derives units similar to place, head-direction, and spatial view cells.   Place cells and head-direction cells are found in rat hippocampus~\citep{o1971hippocampus,taube1990head}, while spatial view cells are found in primate hippocampus~\citep{rolls1999spatial}.

Although BSFA results exhibit the above biological link, it is not clear how this technique might be realized in the brain.  In particular, the space required for a covariance matrix of high-dimensional input is too large.  IncSFA does not require covariance maatrices, and  takes the form of biologically plausible Hebbian and anti-Hebbian updating.

\subsubsection{Hebbian Updating in CCIPCA}  Hebbian updates of synaptic strengths of some neuron make it  more sensitive to expected input activations~\citep{dayan2001theoretical}:

\begin{equation}
\label{EQ:HEBBO}
{\bf v} \leftarrow {\bf v} + \eta~ g({\bf v}, {\bf u}) ~ {\bf u},
\end{equation}

\noindent where ${\bf u}$ represents pre-synaptic (input) activity, and $g$ post-synaptic activity (a function of similarity between synaptic weights ${\bf v}$ and input potentials ${\bf u}$).  The basic Eq.~\ref{EQ:HEBBO} requires additional care (e.g., normalization of ${\bf v})$ to ensure stability during updating.  To handle this in one step, learning rate $\eta$ and retention rate $1-\eta$ can be used,

\begin{equation}
\label{EQ:HEBBO2}
{\bf v} \leftarrow (1-\eta) {\bf v} + \eta~ g({\bf v}, {\bf u}) ~ {\bf u}.
\end{equation}

\noindent where $0 \le \eta \le 1$.  With this formulation, Eq.~\ref{EQ:HEBBIANUPDATING} is Hebbian, where the post-synaptic activity is the normalized response $\displaystyle g({\bf v},{\bf u}) = \frac{{\bf u}_i(t) \cdot {\bf v}_i(t-1) }{ \| {\bf v}_i(t-1) \| }$ and the presynaptic activity is the input ${\bf u}_i$.

\subsubsection{Anti-Hebbian Updating in Peng's MCA}  The general form of anti-Hebbian updating simply results from flipping the sign in Eq.~\ref{EQ:HEBBO}.  In IncSFA notation:

\begin{equation}
\label{EQ:ANTIHEBBO}
{\bf w} \leftarrow {\bf w} - \eta~ g({\bf w}, \dot{{\bf z}}) ~ \dot{{\bf z}}.
\end{equation}

To see the link between Peng's MCA updating and the anti-Hebbian form, in the case of the first MC, we note  Eq.~\ref{EQ:MCAUPDATE} can be rewritten as

\begin{eqnarray}
\label{EQ:ANTIHEBB}
{\bf w}_1 &\leftarrow& 1.5 {\bf w}_1 - \eta ~ \left[ {\bf C}_1 ~ {\bf w}_1 + [{\bf w}_1^T {\bf w}_1] ~ {\bf w}_1 \right], \\
          &\leftarrow& 1.5 {\bf w}_1 - \eta ~ \left[ (\dot{{\bf z}} \cdot {\bf w}_1) ~ \dot{{\bf z}} + ({\bf w}_1 \cdot {\bf w}_1) ~ {\bf w}_1 \right], \\
          &\leftarrow& 1.5 {\bf w}_1 - \eta ~ \|{\bf w}_1\|^2~{\bf w}_1 - \eta ~\left( (\dot{{\bf z}} \cdot {\bf w}_1) ~ \dot{{\bf z}} \right), \\
         &\leftarrow& \left(1.5 - \eta ~ \|{\bf w}_1\|^2\right) ~ {\bf w}_1 - \eta ~ (\dot{{\bf z}} \cdot {\bf w}_1) ~ \dot{{\bf z}},
\end{eqnarray}

\noindent where $(\dot{{\bf z}} \cdot {\bf w}_1)$ indicates post-synaptic strength, and $\dot{{\bf z}}$ pre-synaptic strength.

%The somewhat complicated first term $\left(1.5 - \eta ~ \|{\bf w}_1\|^2\right)$ is needed to control the magnitude of ${\bf w}_1$.

When dealing with nonstationary input, as we do in IncSFA due to the simultaneously learning CCIPCA components, it is acceptable\footnote{Peng: personal communication.} to normalize the magnitude of the slow feature vectors: ${\bf w}_i \leftarrow {\bf w}_i / \|{\bf w}_i\|$.  Normalization ensures non-divergence (see Section~\ref{SE:ANALYSIS}). If we normalize, Eq.~\ref{EQ:ANTIHEBB} can be rewritten in the even simpler form

\begin{eqnarray}
\label{EQ:SIMPLEANTIHEBB}
{\bf w}_1 &\leftarrow& (1 - \eta){\bf w}_1 - \eta (\dot{{\bf z}} \cdot {\bf w}_1) ~ \dot{{\bf z}}, \\
{\bf w}_1 &\leftarrow& {\bf w}_1 / \|{\bf w}_1\|
\end{eqnarray}

\noindent an even more basic anti-Hebbian updating with retention rate and learning rate.  Now, for all other slow features $i>1$, the update can be written so sequential addition shows itself to be a lateral competition term:

\begin{eqnarray}
\label{EQ:LATERAL}
{\bf w}_i &\leftarrow& (1 - \eta){\bf w}_i - \eta \left((\dot{{\bf z}} \cdot {\bf w}_i) ~ \dot{{\bf z}} + \gamma ~ \sum_j^{i-1} ({\bf w}_j \cdot {\bf w}_i) {\bf w}_j \right).
\end{eqnarray}

%Given $\|{\bf w}_i\| = 1$

\section{IncSFA Algorithm}
\label{SE:ALGO}

Now we can present the algorithm for a single IncSFA unit.  For each time step $t=0,1,\ldots$:

\begin{enumerate}
\item \textbf{Sense:} Grab the current raw input as vector $\breve{{\bf x}}(t)$.

\item \textbf{Non-Linear Expansion:}  (optionally) Generate an expanded signal ${\bf x}(t)$ with $I$ components, e.g. for a quadratic expansion:

    \begin{equation}
    \label{EQ:EXPANSION}
    {\bf x}(t) = [\breve{x}_1(t), ..., \breve{x}_d(t), \breve{x}_1^2(t), \breve{x}_1(t) \breve{x}_2(t),... , \breve{x}_d^2(t)]
    \end{equation}

\item \textbf{Mean Estimation and Subtraction:} The signal must be centered (zero mean).  This can be done incrementally if needed.  If $t=0$, set ${\bf \bar{x}}(t) = {\bf x}(0)$.  Otherwise, update mean vector estimate ${\bf \bar{x}}(t)$:

    \begin{equation}
    \label{EQ:MEANUPDATE}
    {\bf \bar{x}}(t) = (1 - \eta) ~ \bar{{\bf x}}(t-1) + \eta ~ {\bf x}(t).
    \end{equation}

\item \textbf{Variance Estimation and Normalization:} (optionally) The variance of the signal can be normalized.  To do so incrementally, the variance estimates ${\bf \sigma} = (\sigma_1, ..., \sigma_I)$ are updated:

    \begin{equation}
    \label{EQ:VARUPDATE}
    \sigma_i(t) = (1 - \eta) ~ \sigma_i(t-1) + \eta ~ (x_i(t) - \bar{x}_i(t))^2, ~\forall i
    \end{equation}

    \noindent and normalize each component's variance by dividing by the estimate.

   \noindent For the following steps, ${\bf u}(t)$ is the processed signal, which has zero mean and unit variance.

    %\begin{equation}
    %\label{EQ:NOMEAN}
    %{\bf u}(t) \leftarrow ({\bf x}(t) - {\bf \bar{x}}(t)) / \sigma(t).
    %\end{equation}

\item \textbf{CCIPCA:} Update estimates of the most significant $K$ principal components of ${\bf u}$, where $K \le I$:

\begin{enumerate}
\item If $t<K$, initialize ${\bf v}_t(t) = {\bf u}(t)$.
\item Otherwise do for $j=1,2,...,K$:  Let ${\bf u}_1(t) = {\bf u}(t)$; execute CCIPCA equations~\ref{EQ:HEBBIANUPDATING} and~\ref{EQ:Residual}.
\end{enumerate}

%order these by eigenvalue

\item \textbf{Whitening and Dimensionality Reduction:} Let ${\bf V}(t)$ contain the normed estimates of the $K$ principal components, ordered by estimated eigenvalue, and create diagonal matrix ${\bf D}(t)$, where $D_{i,i} = 1/\sqrt{\lambda_i(t)}, \forall i \le K$.  Then, ${\bf z}(t) = {\bf V}(t){\bf D}(t){\bf u}(t)$.

\item \textbf{Derivative Signal:} As a forward difference approximation of the derivative, let ${\bf \dot{z}}(t) = {\bf z}(t) - {\bf z}(t-1)$.

\item \textbf{Extract First Principal Component:} Use CCIPCA to update the first PC of ${\bf \dot{z}}$ (to set sequential addition parameter $\gamma(t))$.

\item \textbf{Update Slowness Measure:} The slowness measure of the signal ${\bf \dot{z}}$ is computed and updated incrementally to automatically set the learning rate for MCA.

\item \textbf{Slow Features:} Update estimates of the least significant $J$ PCs of ${\bf \dot{z}}$, where $J \le K$:

    \begin{enumerate}
    \item If $t < J$, initialize ${\bf w}_t = {\bf \dot{z}}(t)$.
    \item Otherwise, let ${\bf C}_1(t) = {\bf \dot{z}}(t) {\bf \dot{z}}^T(t)$, and for each $i=1,...,J$, execute incremental MCA updates in equation~\ref{EQ:LATERAL}.
    \end{enumerate}

\item \textbf{Normalize Slow Feature Estimates:} (optionally, %early on,
for stability) Each ${\bf w}_i \leftarrow {\bf w}_i / \| {\bf w}_i \|.$

\item \textbf{Output:} ${\bf y}(t) = {\bf z}^T(t) {\bf W}(t)$ is the SFA output.

\end{enumerate}

%this has been reduced from section to subsection
\subsection{Convergence of IncSFA}
\label{SE:ANALYSIS}

It is clear that if whitened signal ${\bf z}$ is drawn from a stationary distribution, the MCA convergence proof~\citep{peng2007convergence} applies.  But typically the whitening matrix is being learned simultaneously.  In this early stage, while the CCIPCA vectors are learning, care must be taken to ensure that the slow feature estimates will not diverge.

It was shown that for any initial vector ${\bf w}(0)$ within the set $\mathcal{S}$,

\begin{equation}
\label{EQ:INVSET}
\mathcal{S} = \left\{ {\bf w}(t) | {\bf w}(t) \in \mathcal{R}^K~\mbox{and}~ \|{\bf w}(t)\|^2 \le \displaystyle \frac{1}{2\eta}\right\},
\end{equation}

\noindent will remain in $\mathcal{S}$ throughout the dynamics of the MCA updating.  $\|{\bf w}\|$ must be prevented from getting too large until the whitening matrix is close to accurate.  With respect to lower-order slow features, there is additional dependence on the sequential addition technique, parameterized by $\gamma(t) = \lambda_1(t) + \epsilon$.  This $\gamma(t)$ also needs time to estimate a close value to the first eigenvalue $\lambda_1$. Before these estimates become reasonably accurate, the input can knock the vector out of $\mathcal{S}$.

In practice, normalization of ${\bf w}$ after each update was found to be the most useful.  If $\|{\bf w}(0)\| = 1$ then any learning rate $\eta \le 0.5$ ensures non-divergence.
%However, the convergence is not guaranteed with normalization, and so it is advisable to not do this once the CCIPCA has stabilized.
Another applicable tactic is clipping.  If the signal ${\bf z}$ is thresholded, e.g., from -5 to 5, the potential effect of outliers is controlled.  A third tactic is to use a gradually increasing MCA learning rate.

Even if ${\bf w}$ remains in $\mathcal{S}$, the additional constraint ${\bf w}^T(0){\bf w}^* \ne 0$ is needed for the convergence proof. But this is an easy condition to meet, as it is unlikely that any ${\bf w}(t)$ will be exactly orthogonal to the true feature.  In practice, it may be advisable to add a small amount of noise to the MCA update.  But we did not find this to be necessary.

As for CCIPCA: If the standard conditions on learning rate~\citep{papoulis1965probability} (including convergence at zero), the first stage components will converge to the true PCs, leading to a ``nearly-correct'' whitening matrix in reasonable time.  So, if ${\bf x}$ is stationary, the slow feature estimates are likely to become quite close to the true slow features in a reasonable amount of updates.

In open-ended learning, convergence is not desired.  Yet by using a learning rate that is always nonzero, the stability of the algorithm is reduced.  This corresponds to the well-known stability-plasticity dilemma~\citep{grossberg1980does}.

\subsection{Setting Learning Rates}
\label{sec:AlgParams}

In CCIPCA, if $\eta = \frac{1}{t}$, Eq.~\ref{EQ:HEBBIANUPDATING} will  be the most efficient estimator\footnote{The most efficient estimator on average requires the least samples for learning among all unbiased estimators.  The sample mean is the maximum likelihood estimator (i.e., most efficient unbiased estimator) of the population mean for several distribution types, e.g., Gaussian.} of the principal component.  But a learning rate of $1/t$ is spatiotemporally optimal if every sample from $t=1,2,...,\infty$ is drawn from the same distribution, which will not be the case for the lower-order components, and in general for autonomous agents.  We use an amnesic averaging technique, where the weights of old samples diminuish over time.  Amnesic averages remain unbiased estimators of the true PCs.  For Eq.~\ref{EQ:HEBBIANUPDATING}, $\mbox{E}[{\bf v}(n)] \rightarrow \mbox{E}[{\bf u}]$, as $n \rightarrow \infty$.

To set the CCIPCA learning rate, (and other learning rates, e.g.,  for the input average $\bar{{\bf x}}$), we used the following three-sectioned amnesic averaging function $\mu$:

\begin{equation}
\label{EQ:rho(n)}
 \mu(t) = \left \{
\begin{array}{ll}
0 &  \mbox{if $t \leq t_1,$} \\
c(t - t_1)/(t_2 - t_1) & \mbox{if $t_1 < t\leq t_2,$} \\
c+(t - t_2)/ r & \mbox{if $t_2 < t.$}
\end{array}
 \right.
\end{equation}

Eq.~\ref{EQ:rho(n)} combines optimal updating and plasticity for each feature.  It uses three stages, defined by points $t_1$ and $t_2$.  In the first stage, the learning rate is $\frac{1}{t}$.  In the second, the learning rate is scaled by $c$ to speed up learning of lower-order components.  In the third, it changes with $t$, eventually converging to $1/r$.

Unlike with Peng's MCA, there is no convergence proof for CCIPCA and this type of learning.  Instead, plasticity introduces an expected error that will not vanish~\citep{weng2006optimal}.  To see this, note that any component estimate is a weighted sum of all the inputs:

\begin{equation}
\label{EQ:VTIME}
{\bf v}(t) = \sum_{\tau=1}^t \rho(t) {\bf u}(t),
\end{equation}

\noindent where $\sum_{\tau=1}^t \rho(t) = 1$.  Then,

\begin{equation}
\label{EQ:CCIPCAConvergence} E \|{\bf v}(t) - {\bf v}^* \|^2 =
\displaystyle \sum_{\tau=1}^t \rho^2(t) E\|{\bf u}\|^2 = \displaystyle
\sum_{t=1}^T \rho^2(t) \; tr( E\| {\bf u}{\bf u}^T \|)
\end{equation}

\noindent gives expected estimation error as a function of number of samples
$T$.  Eq.~\ref{EQ:CCIPCAConvergence} can be used to estimate the number of samples needed to get below an acceptable expected error bound, if the signal is stationary.  Otherwise the process retains the ability to adapt at any future $t$.  This introduces some expected error that is linked to the learning rate into the IncSFA whitening process.  Our results show that this is not problematic for many applications, but merely leads to a slight oscillatory behavior around the true features.

To prevent divergence while CCIPCA is still learning, we used a slowly rising learning rate for MCA, starting from low $\eta_l$ at $t=0$ and rising to high $\eta_h$ at $t=T$,

% Required? This plot looks pretty bad
% \begin{figure}[!ht]
% \begin{center}
% \includegraphics[width=8cm]{Figs/eps_T}
% \end{center}
% \caption{$\eta$ vs t}
% \label{fig:eta_T}
% \end{figure}

\begin{equation}
\label{EQ:eta_T}
 \eta(t) = \left \{
\begin{array}{ll}
\eta_l + (\eta_h - \eta_l) * \left(\frac{t}{T}\right)^2 &  \mbox{if $t \leq T,$} \\
\eta_h & \mbox{if $T < t.$}
\end{array}
 \right.
\end{equation}

%\noindent shown visually in Fig.~\ref{fig:eta_T}.

\noindent Ideally, $T$ is a point in time when whitening has stabilized.

%The eventually constant learning rate $\eta_h$ controls how quickly the slow features learn.

The upper bound $\eta_b$ of permissible $\eta_h$ is defined by the first condition in Eq.~\ref{EQ:CONVERGECOND}:

\begin{equation}
\label{EQ:MCALR}
\eta_h < \eta_b = \frac{1}{2 \lambda_1},
\end{equation}

\noindent where $\lambda_1$ is the greatest eigenvalue of the signal. Constant values close to but below the bound can be used to achieve faster convergence.

The algorithm maintains an incremental estimate of intermediate output slowness.  This can be used to automatically adapt the MCA learning rate to changing statistics of the input stream. Since MCA receives a derivative of the whitened input signal, the greatest eigenvalue $\lambda_1$ corresponds to the component that changes most rapidly. As a fast approximation, we set

\begin{equation}
\label{EQ:Lambda1Approx}
\lambda_1 \approx \max_i \Delta(\dot{z}_i) = \Delta(\dot{z}_m),
\end{equation}

\noindent where $z_m$ is the $m^{th}$ dimension of $\bf z$, which has maximal temporal variation.  The $\Delta$-value (\ref{EQ:DELTA}) measures temporal variation of the signal ${\bf x}(t)$. It is given by the mean square of that signal's temporal derivative.  The smaller the $\Delta$-value, the slower the variation of the corresponding signal component.

\begin{equation}
\label{EQ:DELTA}
\Delta({\bf x}) = \avt{\dot{\bf x}(t)^2  }
\end{equation}

The $\Delta$-value is related to Wiskott \& Sejnowski's~\citep{WisSej2002} \textit{slowness measure} of the input signal given by

\begin{equation}
\label{EQ:S}
S({\bf x}) = \displaystyle \frac{P}{2\pi}\sqrt{\Delta ({\bf x}) }
\end{equation}

The value $S$ for some signal of length P indicates how often a pure sine wave of the same $\Delta$ value would oscillate.

Now, from Eq.~\ref{EQ:Lambda1Approx} and Eq.~\ref{EQ:S}, we have

\begin{eqnarray}
\Delta({\dot{z}_m}) &\propto& S(\dot{z}_m)^{2}\label{EQ:Delta_S}\\
\lambda_1 &\propto& S(\dot{z}_m)^2\label{EQ:Lambda_S}
\end{eqnarray}

\noindent Since $\eta_b = \frac{1}{2 \lambda_1}$, we get

\begin{equation}
\label{EQ:etab_slowness}
\eta_b \propto S(\dot{z}_m)^{-2}
\end{equation}

\noindent Selecting $\eta_h$ close to $\eta_b$ (see \ref{EQ:MCALR}), we can write
\begin{equation}
\label{EQ:etah_etab}
\eta_h = \eta_b - \psi
\end{equation}
\noindent for some arbitrarily small constant $\psi$.

From Eq.~\ref{EQ:etab_slowness} and Eq.~\ref{EQ:etah_etab} we get

\begin{equation}
\label{EQ:eta_slowness}
\eta_h \propto S(\dot{z}_m)^{-2}
\end{equation}

%juergen: don't understand this:

With a working learning rate $\eta_h$ and the slowness measure estimate for some input, we can automatically adapt $\eta_h$ for a new signal by tracking how its slowness measure changes.

\subsection{Dimensionality Reduction Parameter}

The eigenvectors of ${\bf x}$ associated with the smallest eigenvalues might represent noise dimensions. Instead of passing this  typically useless information to our second phase, the small eigenvalue directions can be discarded.

While whitening the $I$-dimensional input signal, the dimension can be reduced to $K \le I$.  $K$ can be automatically tuned.  A method we found to be successful is to set $K$ such that no more than a certain percentage of the previously estimated total data variance (the denominator below) is lost.  Let $\beta$ be the ratio of total variance to keep (e.g., 0.95), and compute the smallest $K$ such that

\begin{equation}
\label{EQ:DIMRED}
\displaystyle \frac{ \sum_k^K \lambda_k(t)}{ \sum_i^I \lambda_k(t-1) } > \beta.
\end{equation}

%\section{Hierarchical Processing with IncSFA}
%\label{SE:HARCH}

%\subsection{AutoIncSFA}

%\begin{figure}[!ht]
%\begin{center}
%\includegraphics[width=12cm]{autoincsfa_arch}
%\end{center}
%\caption{Example AutoIncSFA Architecture.}
%\label{fig:autoincsfaarch}
%\end{figure}

%Another strategy for reducing the search space of IncSFA is to spatially compress the input. We found autoencoder (AE) networks to be useful preprocessors for this purpose. Figure \ref{fig:autoincsfaarch} shows an example architecture of an AutoIncSFA network consisting of an AE with 100 hidden units and a single IncSFA unit on top.

%Additional aspects should be taken into account for AE preprocessing.  AEs handle non-linearity quite well and help to avoid the quadratic expansion in the input space.
%An AE with a reduced hidden representation codes only for
%the dominant spatial information in the data, therefore the
%inal output is not severely affected by insignificant yet slowly
%changing environmental elements. Spatial compression also
%helps to eliminate much of the redundant static information
%in the data, resulting in a reduction of the number of IncSFA
%units. A more detailed description along with several real-video experimental results can be found in %\citep{AutoIncSFA2011}.

\section{Experiments and Results}
\label{SE:RESULTS}

Some of our experiments are designed to show that IncSFA derives the same features as batch SFA.  Others show how IncSFA can work in scenarios where batch SFA is not applicable, and how IncSFA can be utilized in high-dimensional video processing applications.   Experiments were done either using Python (using the MDP toolbox~\citep{MDP2008}) or Matlab.

%I guess we can make the code available is accepted?
%This code is available on the first author's website\footnote{\url{www.idsia.ch/~kompella}}.  Others were done using Matlab; this code is available on the second author's site\footnote{\url{www.idsia.ch/~luciw}}.

\subsection{Proof of Concept}

%The first three experiments are designed to show how the single-layer IncSFA compares to single-layer BSFA in settings where the BSFA features are indicative of ground truth.

\begin{figure}[!ht]
\hspace{-0.3in}
\includegraphics[width=15cm]{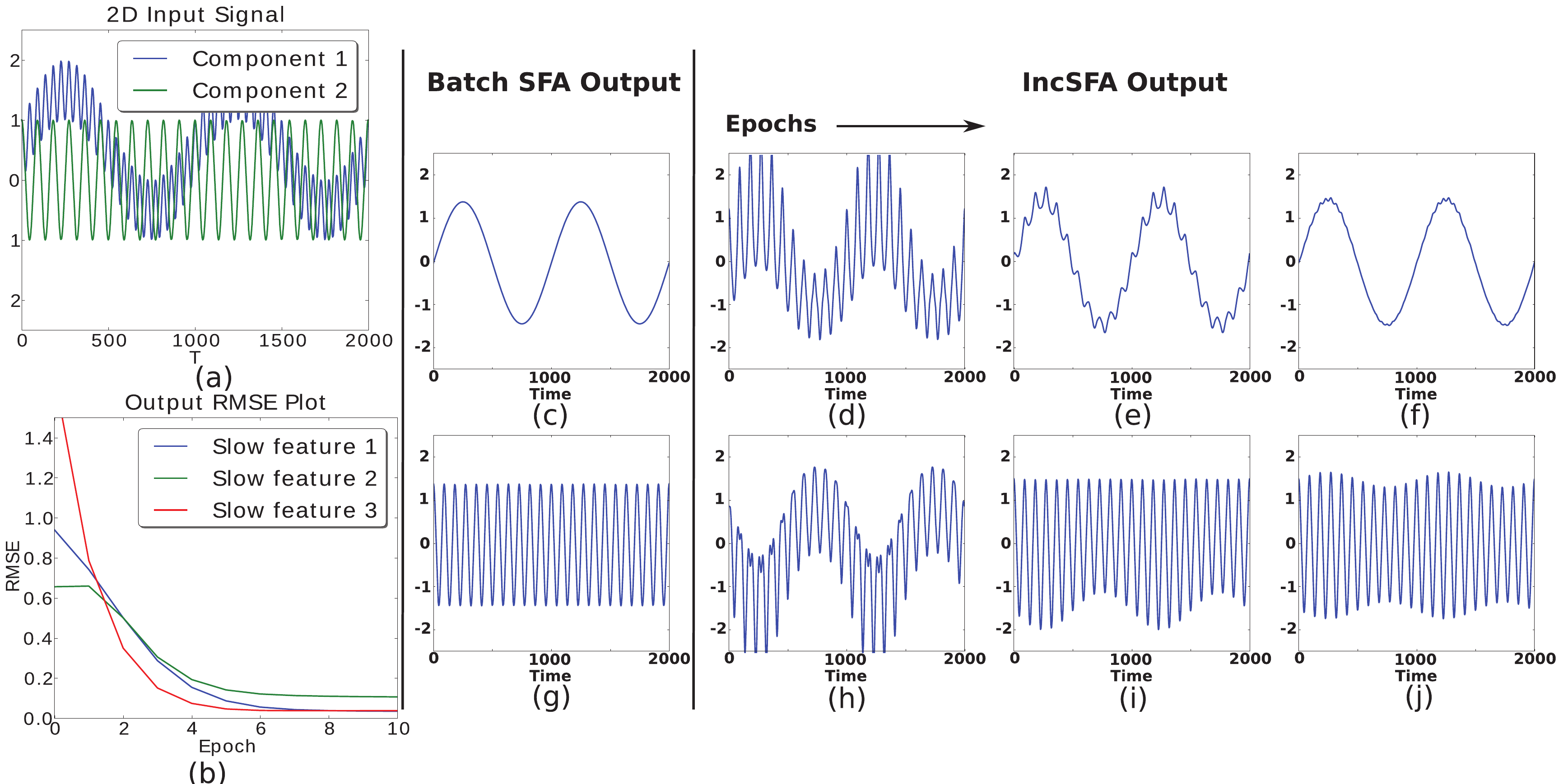}
\caption{Experiment with a simple non-linear input signal. A learning rate of $\eta = 0.08$ is used. (a) Input Signal (b) Output RMSE plot (c) Batch SFA output of the first slow feature (d)-(f) IncSFA output at t = 2, 5, 10 epochs. (g) Batch SFA output of the second slow feature (h)-(j) IncSFA output at t = 2, 5, 10 epochs.}
\label{fig:experiment1}
\end{figure}

As a basic proof of concept, IncSFA is applied to problem introduced in the original SFA paper~\citep{WisSej2002}.  The input signal is

\begin{eqnarray}
\label{EQ:INCSFA1}
\breve{x}_1(t) &=& \mbox{sin}(t) + \mbox{cos}(11~t)^{2}, \\
\breve{x}_2(t) &=& \mbox{cos}(11~t),~t \in [0,2 \pi],
\end{eqnarray}

Both vary quickly over time (see Figure \ref{fig:experiment1}(a)). A total of $2,000$ discrete datapoints are used for learning. The slowest feature hidden in the signal is $y_1(t) = \breve{x}_1(t) - \breve{x}_2(t)^2 = \mbox{sin}(t)$, and the second is $\breve{x}_2(t)^2$.

Both BSFA and IncSFA extract these features.  Figure \ref{fig:experiment1}(b) shows the Root Mean Square Error (\textbf{RMSE}) of IncSFA signals compared to the BSFA output, over multiple epochs of training.  The RMSE at the end of 10 epochs is found to be equal to $[0.0360, 0.1078, 0.0377]^T$.

Figure \ref{fig:experiment1}(c) and (g) shows feature outputs of batch SFA, and (to the right) IncSFA outputs at $2,~5$, and $10$ epochs. Figures \ref{fig:experiment1}(g)-(j) show this comparison for the second feature.

This result show that it is indeed possible to extract multiple slow features in an online way without storing covariance matrices.

\subsection{Extraction of a Driving Force from High Dimensional Input}

\begin{figure}[!ht]
\hspace{-0.3in}
\includegraphics[width=15cm]{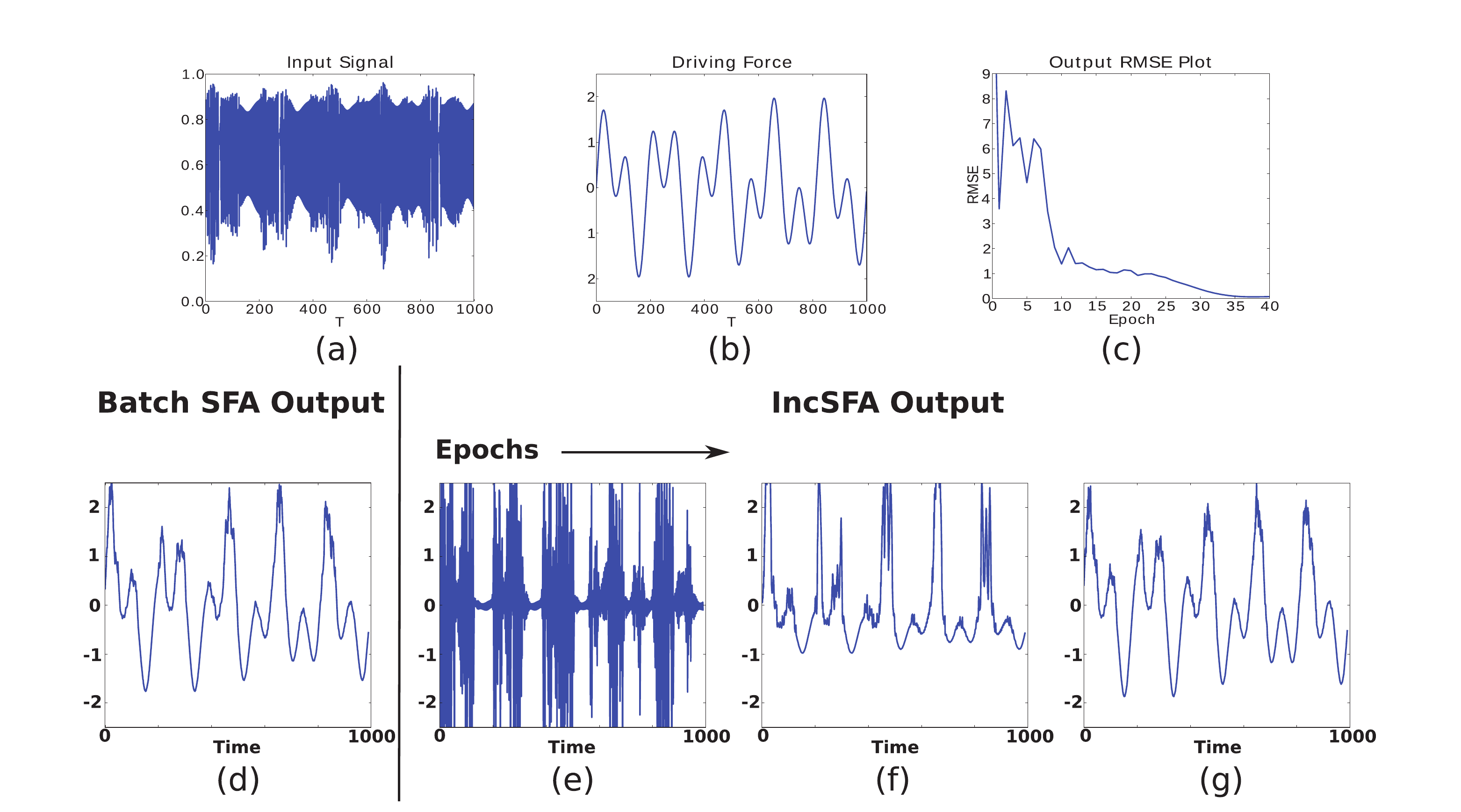}
\caption{ Experiment with a chaotic time series derived from a logistic map. A learning rate of $\eta = 0.004$ is used. (a) Driving Force (b) Input (c) Output RMSE plot (d) BSFA output of the slowest feature (e)-(g) IncSFA output at t = 15, 30, 60 epochs. }
\label{fig:experiment_2}
\end{figure}

A classic slow feature extraction problem involves uncovering the driving force of a dynamic system hidden in a very complex signal. Here, a chaotic time series is derived from a logistic map~\citep{MDP2008}:

\begin{equation}
\breve{x}(t+1) = (3.6 + 0.13 ~ \gamma(t)) \breve{x}(t) ~ (1 - x(t)),
\end{equation}

\noindent which is driven by a slowly varying driving force $\gamma(t)$ made up of two frequency components (5 and 11 Hz) given by

\begin{equation}
\label{eqn:drvngfrc}
\gamma(t) = \mbox{sin}(10 \pi t) + \mbox{sin}(22 \pi t).
\end{equation}

To show the complexity of the signal, figures \ref{fig:experiment_2}(a) and \ref{fig:experiment_2}(b) plot the driving force signal $\gamma(t)$ and the generated time series $\breve{x}(t)$, respectively.

A total of $1,000$ discrete datapoints are used.  The driving force cannot be extracted linearly, so a nonlinear expansion is used---temporal in this case.  The signal is embedded in 10 dimensional space using a sliding temporal window of size 10 (the \textit{TimeFramesNode} from the MDP toolkit \citep{MDP2008} is used for this).  The signal is then spatially  quadratically expanded to generate an input signal with 65 dimensions.

Figure \ref{fig:experiment_2}(c) shows the convergence of IncSFA on the BSFA output,  Figure \ref{fig:experiment_2}(d) BSFA output, and Figures \ref{fig:experiment_2}(e)-(g) the outputs of IncSFA at 15, 30 and 60 epochs.  The \textbf{RMSE} at $60^{th}$ epoch is found to be equal to $0.0984$.

\subsection{Invariant Spatial Coding from Simple Movement Data}
\label{exp:Agentexp}

\begin{figure}[!ht]
\centering{
\hspace{-0.2cm}
\begin{minipage}[b]{1.0\linewidth}
\includegraphics[width=12.8cm]{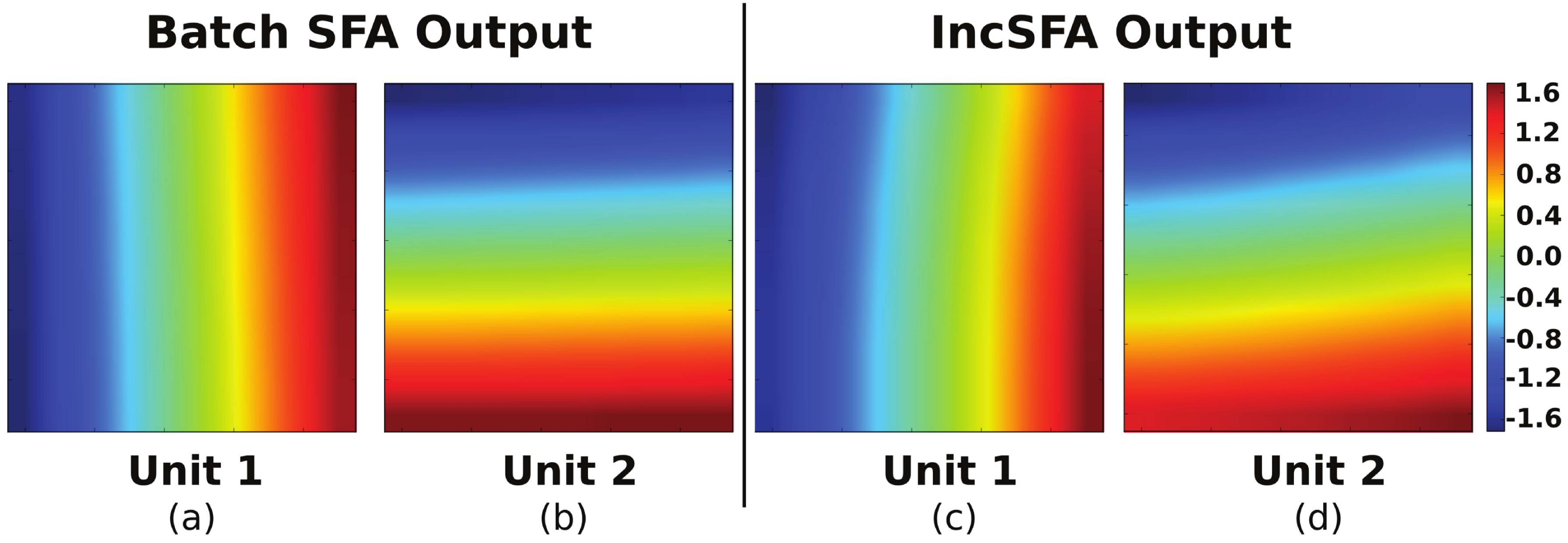}
\end{minipage}
\caption{(a) BSFA output of the first slow feature and (b) the second slow feature (c) IncSFA output of the first slow feature and (d) the second slow feature after 50,000 samples with  learning rate $\eta = 0.003$ (figures best viewed in color).}
\label{fig:experiment_3}}
\end{figure}

%Franzius \textit{et al.}~\citep{franzius2007slowness}, who fed SFA into both ICA or a competitive learning layer to derive computational \textit{place cells} active at specific agent locations, like those  in the rat hippocampus.  Such invariant representations of the environment are of course useful for navigational tasks of organisms or robots.

%Epochs over the same sequence are not needed; IncSFA also works if data is generated from some \textit{movement paradigm} which may not lead to exact sequence repetitions.

Our simulated agent performs a random walk in a two-dimensional bounded space.  Brownian motion is used to generate agent trajectories approximately like those of rats. The agent's position $p(t) = [x(t), y(t)]$ is updated by a weighted sum of the current velocity and gaussian white noise, with standard deviation $v_r$.  The momentum term $m$ can assume values between zero and one, so that higher values of $m$ lead to smoother trajectories and more homogeneous sampling of space in less time. Once the agent is predicted to cross the spatial boundaries, the current velocity is halved and an alternative random velocity update is generated, until a new valid position is reached.  Noise variance $v_{r} = [3.0, 2.5]^T$, mass $m=0.75$ and  $50,000$ data points are used for generating the training set.  A separate test grid dataset samples positions and orientations at regular intervals, and is used for evaluation.

Here is the used movement paradigm:\\
\noindent$~~currVel \leftarrow p(t) - p(t-1);\\
~~\textbf{repeat}\\
~~~~~noise \leftarrow GaussianWhiteNoise2d() * v_r;\\
~~~~~p(t+1) \leftarrow p(t) + m * currVel + (1-m) * noise;\\
~~~~~\textbf{if}~not~isInsideWalkArea(p(t+1)):\\
~~~~~~~~currVel \leftarrow currVel / 2;\\
~~\textbf{until} ~ isInsideWalkArea(p(t+1))\\$
Under this movement paradigm~\citep{franzius2007slowness}, SFA yields slow feature outputs in the form of half-sinusoids, shown in Figure \ref{fig:experiment_3}.  These features collectively encode the agent's $x$ and $y$ position in the environment.  The first slow feature (Figure \ref{fig:experiment_3}(a)) is invariant to the agent's $x$ position, the second  (Figure \ref{fig:experiment_3}(b)) to its $y$ position ($y$ axis horizontal).  IncSFA's results (Figures \ref{fig:experiment_3}(c)-(d)) are close to the ones of the batch version, with an \textbf{RMSE} of $[0.0536,  0.0914]^T$.

\subsection{Feature Adaptation to a Changing Environment}

\begin{figure}[!ht]
\hspace{-0.65cm}
\includegraphics[width=15cm]{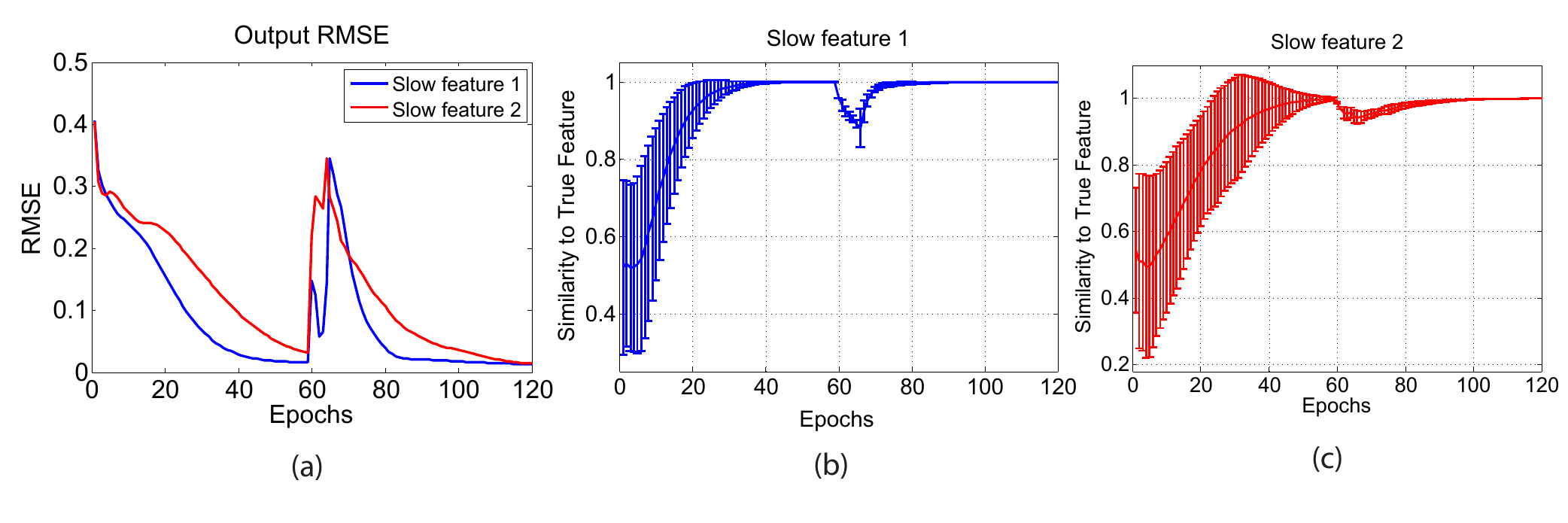}
\caption{(a) RMSE of IncSFA's first two output functions with respect to the true functions for original signal (epochs 1-59), and switched signal (epochs 60-120). (b) Normalized similarity (direction cosine) of the first slow feature to the true first slow feature of the current process, over 25 independent runs.  (c) Normalized similarity of the second incremental slow feature.}
\label{fig:adaptres}
\end{figure}

The purpose of this experiment is to illustrate how IncSFA's features \textit{adapt} to a sudden shift in the input process.  The input used is the same signal as in Experiment \#1, but broken into two partitions.  At epoch 60, the two input lines $x_1$ and $x_2$ are switched such that the $x_1$ signal suddenly carries what $x_2$ used to, and vice versa.  We wish to show that IncSFA can first learn the slow features of the first partition, then is able to adapt to learn the slow features of the second partition.

The signal is sampled 500 times per epoch. The CCIPCA learning rate parameters, also used to set the learning rate of the input average $\bar{{\bf x}}$, were set to $t_1 = 20, t_2 = 200, c=4, r=5000$.  The MCA learning rate is $\eta = 0.01$.

\begin{figure}[!ht]
\hspace{0.1cm}
\includegraphics[width=12cm]{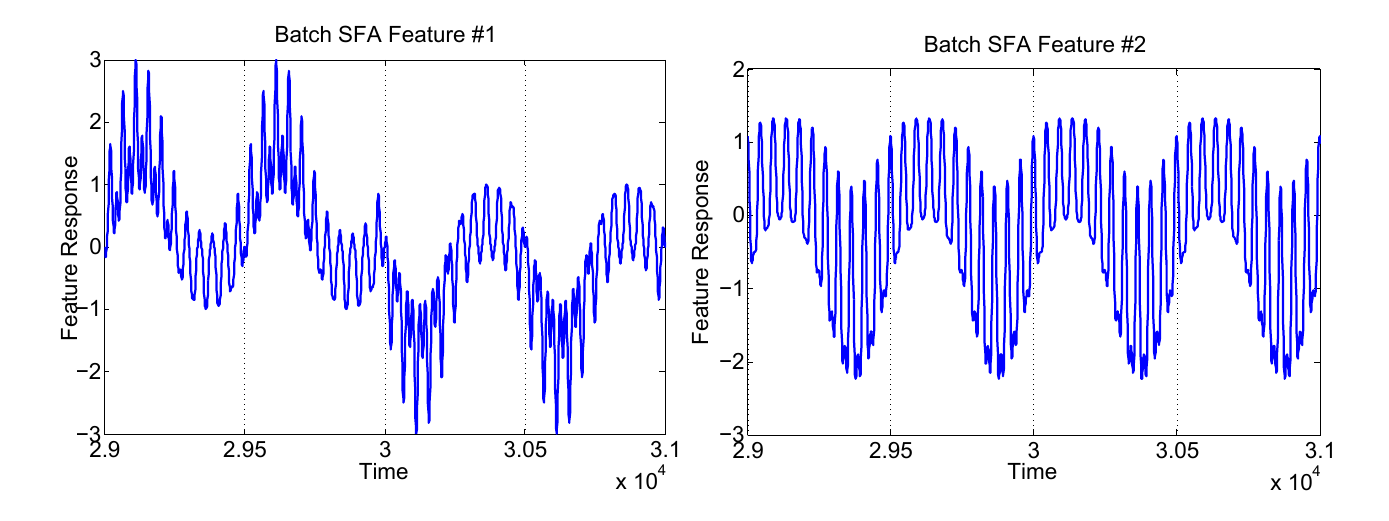}
\caption{Outputs of first two slow features, from epoch 59 through 61, extracted by batch SFA over the input sequence.}
\label{fig:batchadapt}
\end{figure}

Results of IncSFA are shown in Fig.~\ref{fig:adaptres}, demonstrating successful adaptation.  To measure convergence accuracy, we use the direction cosine~\citep{chatterjee2000algorithms} between the estimated feature ${\bf w}(t)$ and true (unit length) feature ${\bf w}^*$,

\begin{equation}
\label{EQ:DCOS}
Direction Cosine(t) = \displaystyle \frac{ | {\bf w}^T(t) \cdot {\bf w}^* | }{ \| {\bf w}^T(t) \| \cdot \|{\bf w}^* \| },
\end{equation}

\noindent  The direction cosine equals one when the directions align (the feature is correct) and zero when they are orthogonal.

BSFA results are shown in Fig.~\ref{fig:batchadapt}.  The first batch feature catches the meta-dynamics and could actually be used to roughly sense the signal switch.  However, the dynamics within each partition are not extracted.

\subsection{Recovery from Outliers}

\begin{figure}[!ht]
%\hspace{in}
\includegraphics[width=12cm]{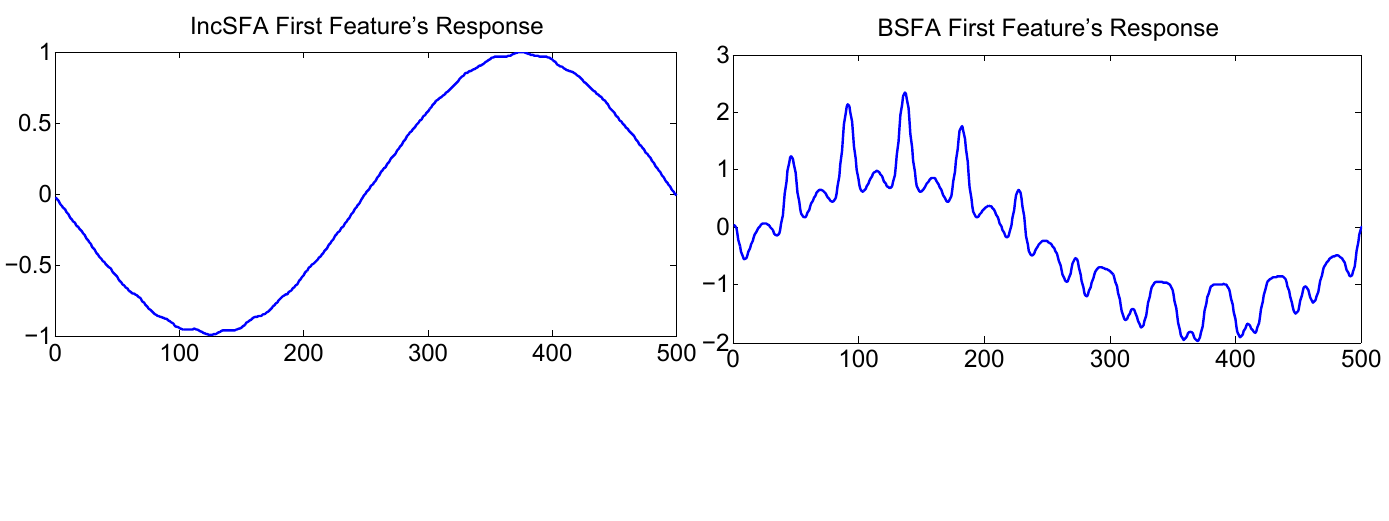}
\vspace{-0.3in}
\caption{First output signals of IncSFA and BSFA on the simple signal with a single outlier.}
\label{fig:outlier}
\end{figure}

Again, the learning rate setup and basic signal from the previous experiment is used, over 150 epochs, with 500 samples per epoch.   A single outlier point is inserted: $x_1(100) = x_2(100) = 2000$.  Figure~\ref{fig:outlier} shows the first output signal of BSFA and IncSFA, showing that the one outlier point at time 100 (out of 75,000) is enough to corrupt the first feature of BSFA, whereas IncSFA recovers.

The relative lack of sensitivity of IncSFA to outliers is shown in a real-world experiment~\citep{AutoIncSFA2011}, in which a person moves back and forth in front of a stable camera.  At only one point in the training sequence, a door in the background is opened, and the BSFA hierarchical network's first slow feature became sensitive to this event.  Yet, the AutoIncSFA network's first slow feature encodes the relative distance of the moving interactor.

\subsection{High-Dimensional Video with Linear IncSFA}

\begin{figure}[!ht]
%\begin{center}
\hspace{-0.3in}
\includegraphics[width=16cm]{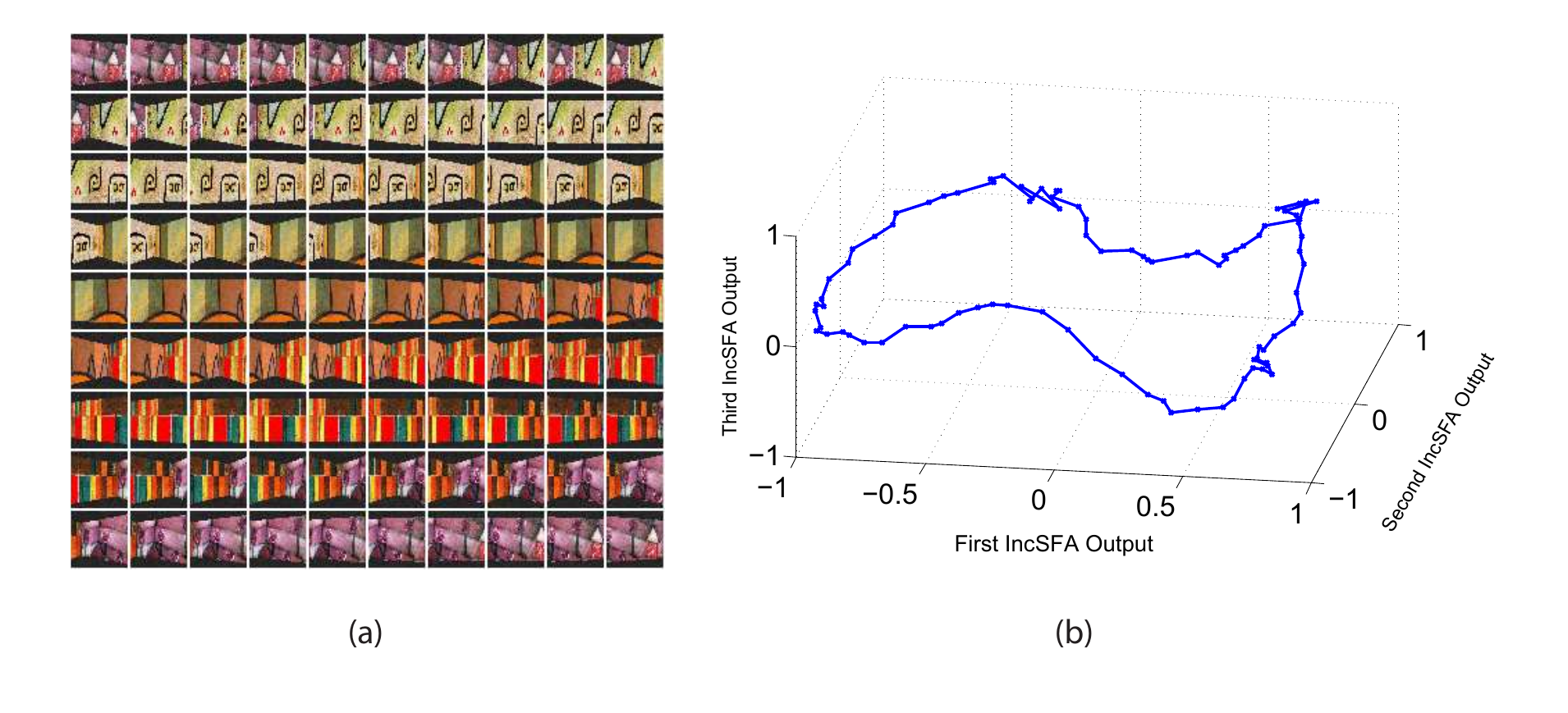}
%\end{center}
\caption{(a) Stream of 90 $41 \times 41 \times 3$ images as the agent  completes one turn (360 degrees). Viewed row-wise, left to right.  (b) Data projected onto the first three features learned by IncSFA.  This gives a compact encoding of the agent's state. }
\label{fig:spinresult}
\end{figure}

IncSFA's scalability is tested with an image sequence of dimension $41 \times 41 \times 3$ (color images: see Fig.~\ref{fig:spinresult}(a)).   The agent is located in the middle of a square room with four complex-textured walls.  In each episode, starting from a different orientation, the agent rotates slowly (4 degree shifts from one image to the next) by 360 degrees.  At any time, a slight amount of Gaussian noise is added to the image ($\sigma = 8$).  The agent has a video input sensor, and the sequence of image frames with $5,043$ dimensions is fed into a linear IncSFA directly. %This experiment tests how well the basic algorithm deals with very high dimensionality.

%(which were also used for the online mean estimation as well as the first eigenvalue in the whitened difference space)

To reduce computation time, only the $40$ most significant principal components are computed by CCIPCA, using learning rate parameters $t_1 = 20,~t_2 = 200,~c=4,~r=5000$.  Computation of the covariance matrix and its full eigendecomposition (including over 5000 eigenvectors and eigenvalues) is avoided.  On the $40 \times 40$ whitened difference space, only the first $5$ slow features are computed via MCA and sequential addition.  $500$ epochs through the data took approximately 15 minutes using Matlab on a machine with an Intel i3 CPU and 4 GB RAM.

The result of projecting the (noise-free) data onto the first three slow features are shown in Fig.~\ref{fig:spinresult}(b).  A single linear IncSFA has incrementally compressed this high-dimensional noisy sequence to a nearly unambiguous compact form, learning to ignore the details at the pixel level and attend to the true cyclical nature underlying the image sequence.  A few subsequences have somewhat ambiguous encodings, because certain images associated with slightly different angles are very similar.

%In any case, this result shows that IncSFA can indeed handle very high-dimensional data.

\subsection{High-Dimensional Video and Episodic Learning}

``Real-world'' learning systems might operate in series of several episodes of interactions with the environment.  IncSFA can be readily extended to episodic tasks, with a minor modification: The derivative signal, which is computed as a difference over a single time step, is simply not computed for the starting sample of each episode.  The first data point in each episode is used for updating the PCs, but not the slow feature vectors.

 \begin{figure}[!h]
 \centering
 \vspace{-0.3in}
 \includegraphics[width=\linewidth]{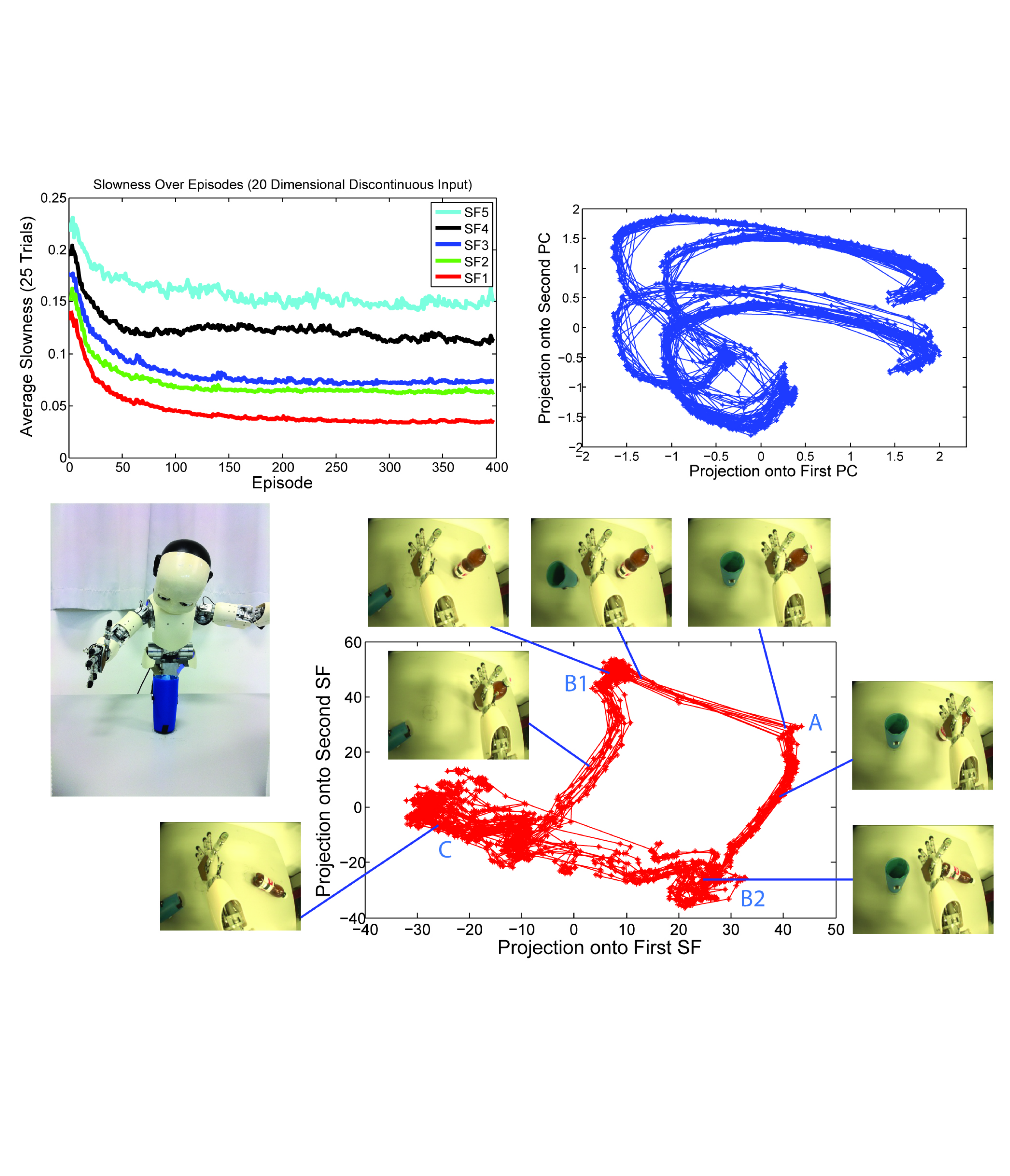}
 \vspace{-1.4in}
 \caption{Experimental result of IncSFA on episodes where the iCub knocks down two cups via motor babbling on one joint.  Upper left: The average slowness of the five features at each episode. Upper right: after training, several episodes (each episode is an image sequence where the cups are eventually both knocked down) are embedded in the space spanned by the first two PCs.  Lower right: the same episodes are embedded in the space spanned by the first two slow features.  We show some example images and where they lie in the embedding.  The cluster in the upper right (A) represents when both cups are upright.  When the robot knocks down the blue cup first, it moves to the cluster in the upper left (B1).  If it instead knocks down the brown cup, it moves to the lower right cluster (B2).  Once it knocks down both cups, it moves to the lower left area (C).}
 \label{fig:single_ul}
 \end{figure}	

\begin{figure}[!h]
\centering
\includegraphics[width=0.5\linewidth]{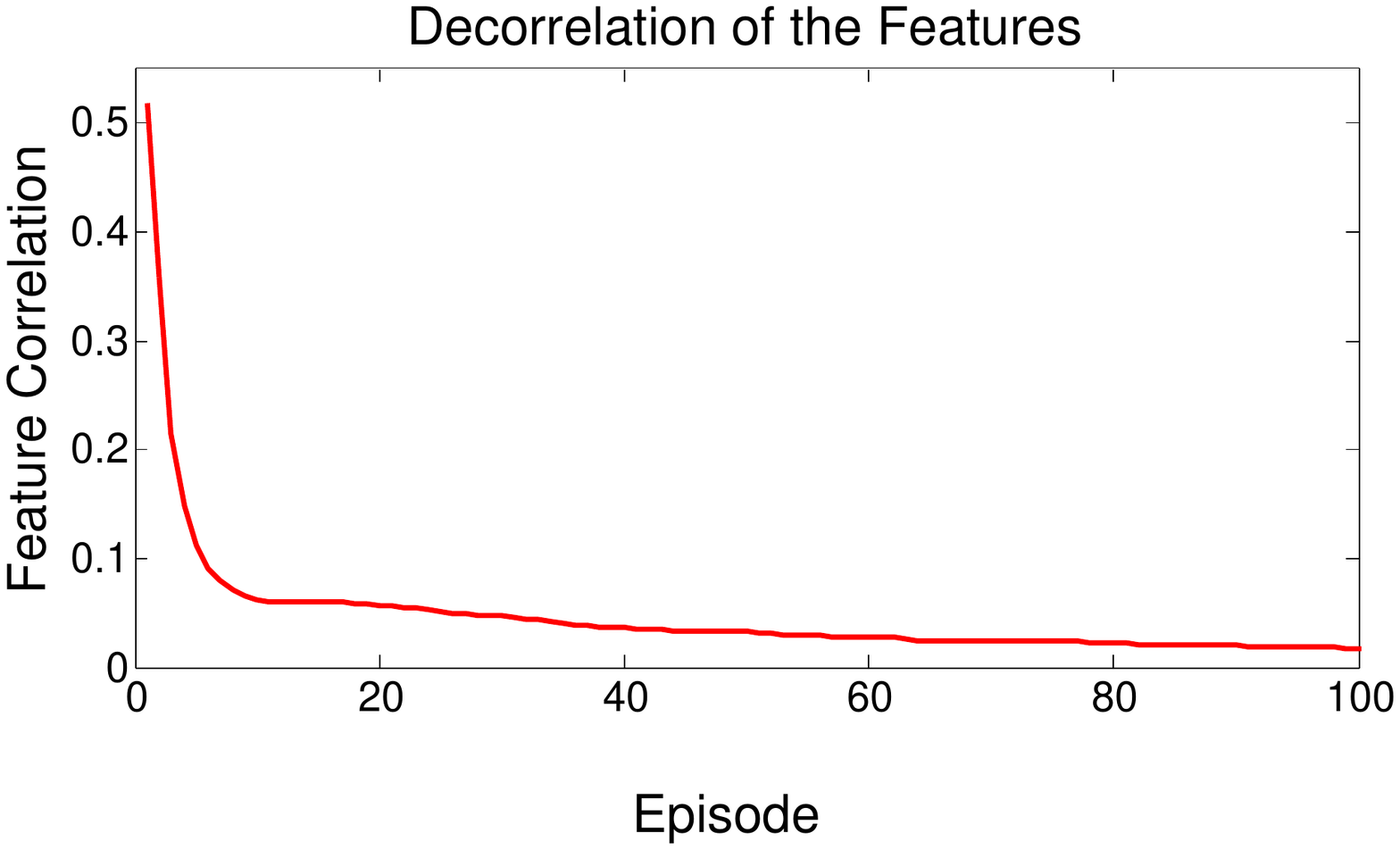}
\vspace{-1.3in}
\caption{Average slow feature similarity over episodes.}
\label{fig:fcorr}
\end{figure}	

%\citep{AutoIncSFA2011}.

Here we present results obtained through a robot's episodic interactions with objects in its field of view.  Two plastic cups are placed in the iCub robot's field of view. The robot performs motor babbling in one joint using a movement paradigm of Franzius \textit{et al.} During the course of babbling, it happens to topple the cups, in one of two possible orders.  The episode ends a short time after it has knocked both down.  A new episode begins with the cups upright again and the arm in the beginning position.  A total of $50$ separate episodes were used as training data.

Linear IncSFA is used on the entire $80 \times 60$ (grayscale) image.  Only the $20$ most significant principal components are computed by CCIPCA, using learning rate parameters $t_1 = 20,~t_2 = 200,~c=2,~r=10000$.  Only the first $5$ slow features are computed via MCA and sequential addition, with learning rate $0.001$.  The MCA vectors are normalized after each update during the first $10$ epochs, but not thereafter (for faster convergence).  Each of 25 different trials was over $400$ randomly-selected (of the 50 possible) episodes.

%took approximately 15 minutes using Matlab on a machine with an Intel i3 CPU and 4 GB RAM.

Results are shown in Fig.~\ref{fig:single_ul}.  We measured the slowness of the features on three ``testing'' episodes, after each episode of training.  The upper left plot shows that all five features get slower over the episodes.  After training completes, we can embed the data in a lower dimension with respect to the learned features.  The embedding of 20 episodes are shown with respect to the first two PCs as well as the first two slow features.  Since the cups being toppled or upright are the slow events in the scene, IncSFA's encoding is keyed on the object's state (toppled or upright).  PCA does not find such an encoding, being much more sensitive to the arm.  Since these events occurs once within each episode, BSFA cannot be used to learn these features.  Figure~\ref{fig:fcorr} shows the average mutual direction cosine between non-identical pairs of slow features, and we can see the features quickly become nearly decorrelated.

Such clear object-specific low-dimensional encoding, invariant to the robot's arm position, is useful, greatly facilitating  training of a subsequent regressor or reinforcement learner. A video of the experimental result can be found at \protect{\url{http://www.idsia.ch/~luciw/IncSFAArm/IncSFAArm.html}}.

\subsection{Hierarchical IncSFA}

\begin{figure}[!ht]
\begin{center}
\includegraphics[width=12cm]{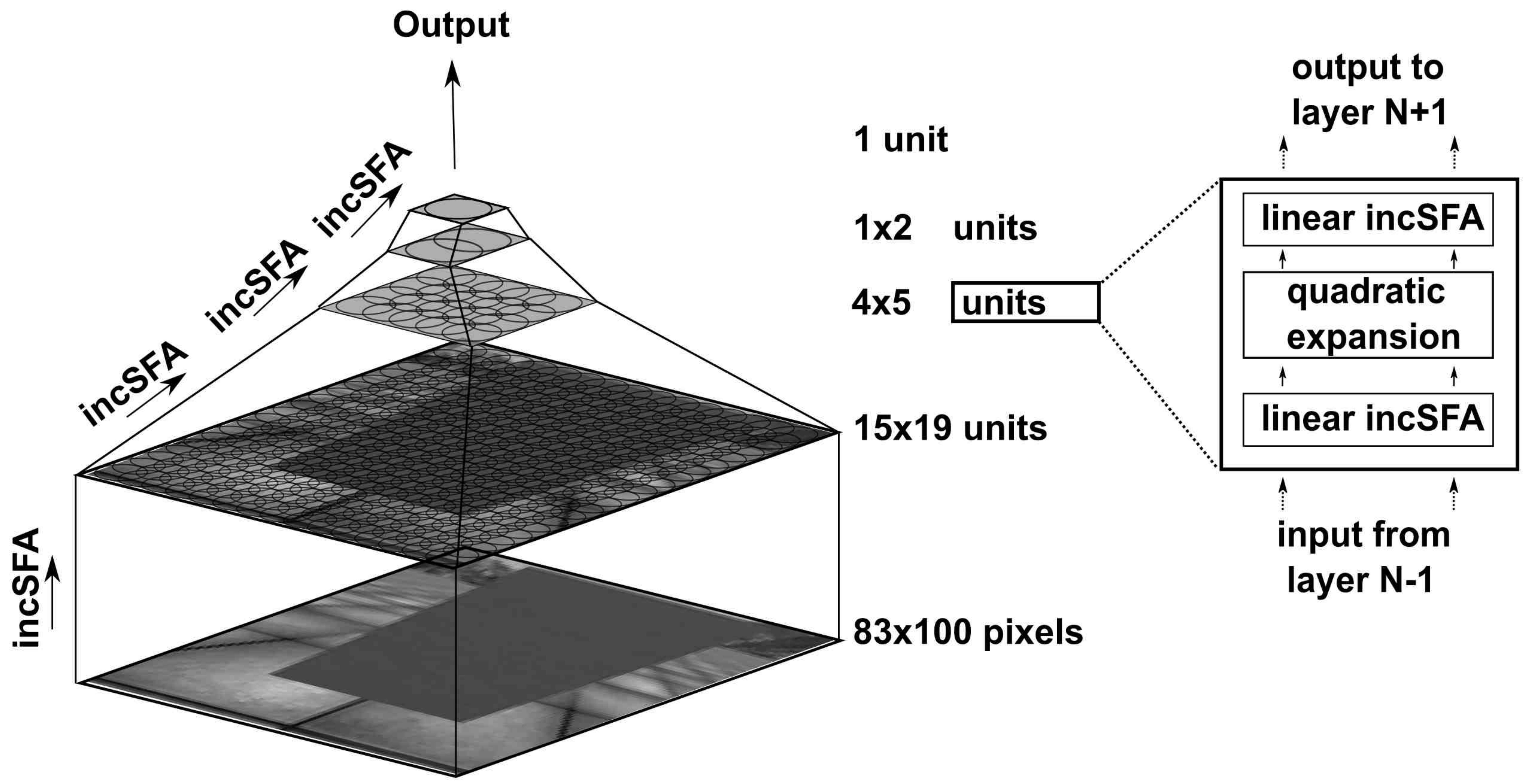}
\end{center}
\caption{Example Hierarchical IncSFA Architecture, also showing the structure of an IncSFA node, which contains a linear IncSFA unit followed by nonlinear expansion followed by another linear IncSFA unit.}
\label{fig:arch}
\end{figure}

Deep networks composed of multiple stacked IncSFA nodes, each sensitive to only a small part of the input (i.e., receptive fields), can be used for processing high-dimensional image streams in a biologically plausible way.  The computational reason for doing this is that very high-dimensional signals correspond to large search spaces, and hierarchical setups breaking up the signal can reduce the search burden.  And using receptive fields reduces the number of necessary lower-order PCs that have to be computed by CCIPCA, which should speed the learning.

Figure \ref{fig:arch} shows an example deep network, motivated by the human visual system and based on the one specified by Franzius \textit{et al.}~\citep{franzius2007slowness}.  The network is made up of a converging hierarchy of layers of IncSFA nodes, with overlapping rectangular receptive fields. Each IncSFA node finds the slowest output features from its input within the subspace of all monomials (e.g., of degree two if a quadratic expansion is used) of the node's inputs.

\begin{figure}[!ht]
\begin{center}
\includegraphics[width=0.95\linewidth]{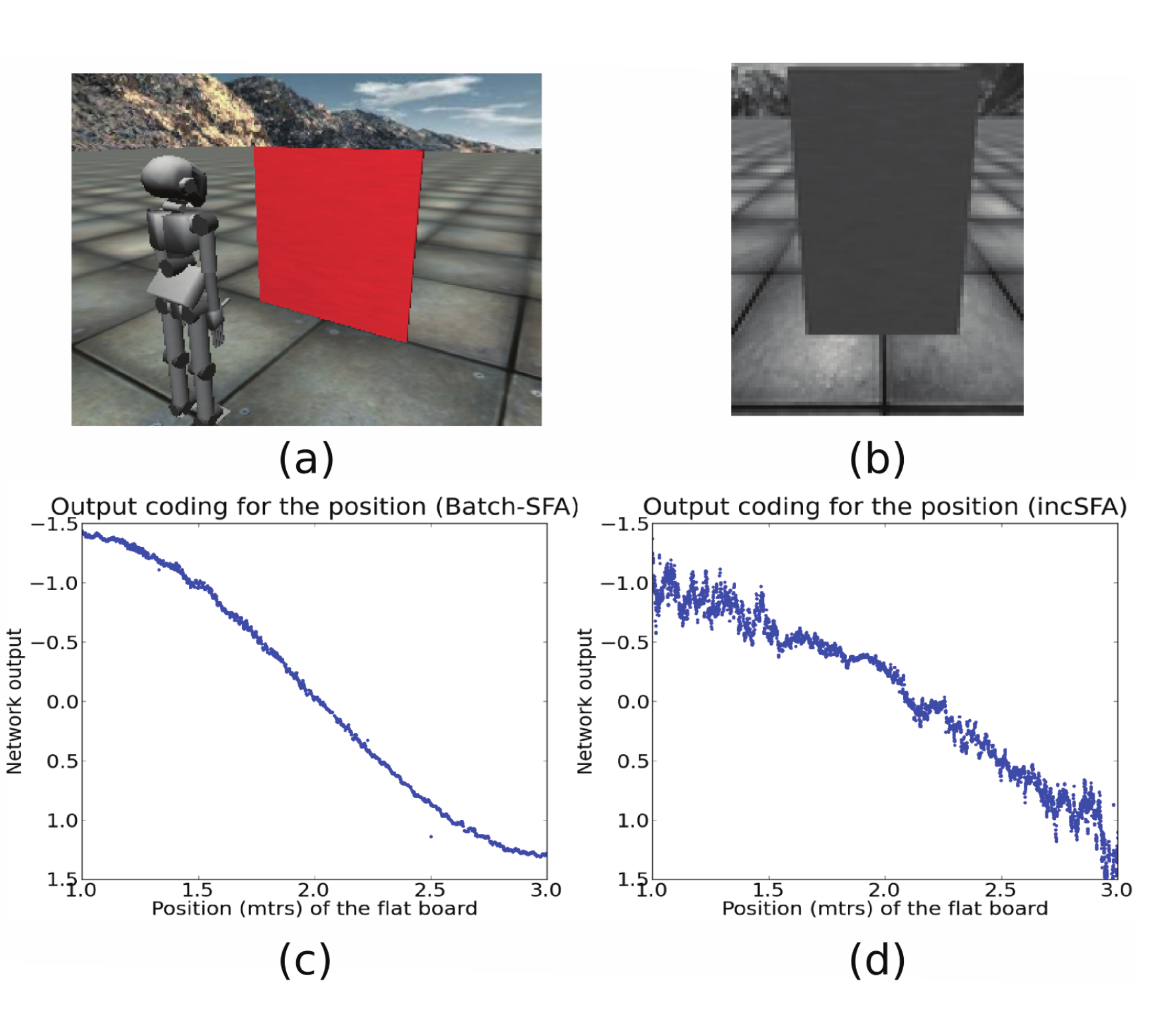}
\end{center}
\caption{(a) Experimental Setup: iCub Simulator (b) Sample image from the input dataset (c) Batch-SFA output (d) IncSFA output ($\eta = 0.005$)}
\label{fig:img_experiment}
\end{figure}

%for this experiment, as a proof of concept with hierarchical architecture
We feed IncSFA with images from a high-dimensional video stream  generated by the iCub simulator~\citep{iCubSim}, an OpenGL-based software specifically built for the iCub robot. Our experiment mimics the robot observing a moving interactor agent, which in the simulation takes the form of a rectangular flat board moving back and forth in depth over the range $[1,3]$ (meters) in front of the robot, using a movement paradigm similar to the one discussed in Section \ref{exp:Agentexp}.  Figure \ref{fig:img_experiment}(a) shows the experimental setup in the iCub simulator. Figure \ref{fig:img_experiment}(b) shows a sample image from the dataset.  $20,000$ monocular images are captured from the robot's left eye and downsampled to 83$\times$100 pixels (input dimension of $8,300$).

%hierarchical network
A three-layer IncSFA network is used to encode the images.  Each SFA node operates on a spatial receptive field of the layer below.  The first layer uses $15 \times 19$ nodes, each with $10 \times 10$ image patch receptive field and a $5$ pixel overlap.  Each node on this layer develops 10 slow features.  The second layer uses $4 \times 5$ nodes, each having a $5 \times 5$ receptive field, and developing 5 slow features.  The third layer uses two nodes, one sensitive to the top half, the other sensitive to the bottom half (5 slow features).  The forth layer uses a single node and a single slow feature.  The network is trained layer-wise from bottom to top, with the lower layers frozen once a new layer begins its training. The CCIPCA output of all nodes is clipped to $[-5 ,5]$,  to avoid any outliers that may arise due to close-to-zero eigenvalues in some of the receptive fields that contain unchanging stimuli. Each IncSFA node is trained individually, that is, there is no weight sharing among nodes.

%An example network used in our experiments breaks the image down as follows: the captured image is sub-sampled from  800 by 600 pixels down to 83 by 100. On the lowest layer, the receptive field of each node consists of a 10 by 10 image patch. The nodes form a regular (i.e., non-foveated) 15 by 19 grid with partially overlapping receptive fields that jointly cover the input image. The second layer with partially overlapping 5 by 5 receptive fields forms a 4 by 5 grid. Similarly, the third and the fourth layer will generate the slowly varying feature-vector output.

%In the case of $83 \times 100$ input images, a network was constructed as follows.  On the lowest layer, the receptive field of each module consists of an image patch of $10 \times 10$ pixels. The output from the first layer forms a $15 \times 19$ grid with 10 slow features per each module. With overlapping receptive fields of $5 \times 5 $, this layer's output, with 5 slow features per module, becomes a $4 \times 5 \times 5$ grid.  Similarly, third and fourth layers decrease signal dimension to produce one single output.

%Given this movement paradigm, the slowest feature at the top of the network will code for distance.  The hierarchical network is needed 1. due to the high image dimensionality and 2. since the function mapping pixel intensities to distance is highly nonlinear.

For comparison, a batch SFA hierarchical network was also trained on this data.  Figures \ref{fig:img_experiment} show BSFA and IncSFA outputs. The expected output is of the form of a sinusoid extending over the range of board positions.  IncSFA gives a slightly noisy output, probably due to the constant dimensionality reduction value for all units in each layer of the network, selected to maintain a consistent input structure for the subsequent layer; hence some units with eigenvectors corresponding to very small eigenvalues emerge in the first stage, with receptive fields observing comparatively few input changes, thus slightly corrupting the whitening result, and adding small fluctuations to the overall result.

Finally, we evaluate how well the IncSFA feature codes for distance.  A supervised quadratic regressor is trained with ground truth labels on 20\% of the dataset, and tested on the other 80\%, to measure the quality of features for some classifier or reinforcement learner using them (see RMSE plot).  Hierarchical IncSFA derives the driving forces from a complex and continuous input video stream in a completely online and unsupervised manner.

\section{Conclusions}
\label{SE:CONCLUDE}

Our novel Incremental Slow Feature Analysis technique solves SFA problems incrementally without storing covariance matrices. IncSFA's covariance-free Hebbian and anti-Hebbian updates add biological plausibility to SFA itself. While batch SFA cannot handle certain open-ended uncontrolled settings, IncSFA can. This makes it a promising tool for learning autonomous robots. Future work will study online learning controllers whose experiments actively create data exhibiting novel but learnable regularities measured by  \textit{improvements} of emerging slow features, in line with the formal theory of curiosity~\citep{schmidhuber2010formal}.

\subsection*{Acknowledgments}

The experimental paradigm used for the distance-encoding high-dimensional video experiment was first developed by the first author under the supervision of Mathias Franzius, at the Honda Research Institute Europe. We would like to acknowledge Dr. Franzius for his contributions in this regard. We would also like to acknowledge IDSIA researchers Alexander Forster and Kail Frank for enabling the experiment with the iCub robot.  Thanks to Marijn Stollenga and Sohrob Kazerounian for their comments on an earlier draft of the paper.  This work was funded through the 7th framework program of the EU under grants \#231722 (IM-Clever project) and \#270247 (NeuralDynamics project), and by Swiss National Science Foundation grant CRSIKO-122697 (Sinergia project).

%Tom Schaul, Leo Pape Jonathan Masci, and Kail Frank for their contributions to the paper on AutoIncSFA.

\label{SE:bib}


\begin{thebibliography}{}

\bibitem[\protect\astroncite{Abut}{1990}]{Abut90}
Abut, H., editor (1990).
\newblock {\em Vector Quantization}.
\newblock IEEE Press, Piscataway, NJ.

\bibitem[\protect\astroncite{Amari}{1977}]{amari1977neural}
Amari, S. (1977).
\newblock Neural theory of association and concept-formation.
\newblock {\em Biological Cybernetics}, 26(3):175--185.

\bibitem[\protect\astroncite{Barlow}{2001}]{barlow2001redundancy}
Barlow, H. (2001).
\newblock Redundancy reduction revisited.
\newblock {\em Network: Computation in Neural Systems}, 12(3):241--253.

\bibitem[\protect\astroncite{Bergstra and Bengio}{2009}]{NIPS2009_0933}
Bergstra, J. and Bengio, Y. (2009).
\newblock Slow, decorrelated features for pretraining complex cell-like
  networks.
\newblock {\em Advances in Neural Information Processing Systems 22}, pages
  99--107.

\bibitem[\protect\astroncite{Chatterjee
  et~al.}{2000}]{chatterjee2000algorithms}
Chatterjee, C., Kang, Z., and Roychowdhury, V. (2000).
\newblock Algorithms for accelerated convergence of adaptive pca.
\newblock {\em IEEE Transactions on Neural Networks}, 11(2):338--355.

\bibitem[\protect\astroncite{Chen et~al.}{1998}]{chen1998unified}
Chen, T., Amari, S., and Lin, Q. (1998).
\newblock A unified algorithm for principal and minor components extraction.
\newblock {\em Neural Networks}, 11(3):385--390.

\bibitem[\protect\astroncite{Chen et~al.}{2001}]{chen2001sequential}
Chen, T., Amari, S., and Murata, N. (2001).
\newblock Sequential extraction of minor components.
\newblock {\em Neural Processing Letters}, 13(3):195--201.

\bibitem[\protect\astroncite{Comon}{1994}]{Comon94}
Comon, P. (1994).
\newblock Independent component analysis, {A} new concept?
\newblock {\em Signal Processing}, 36:287--314.

\bibitem[\protect\astroncite{Dayan and Abbott}{2001}]{dayan2001theoretical}
Dayan, P. and Abbott, L. (2001).
\newblock Theoretical neuroscience: Computational and mathematical modeling of
  neural systems.

\bibitem[\protect\astroncite{Doersch et~al.}{}]{doerschtemporal}
Doersch, C., Lee, T., Huang, G., and Miller, E.
\newblock Temporal continuity learning for convolutional deep belief networks.

\bibitem[\protect\astroncite{F{\"o}ldi{\'a}k}{1991}]{földiák1991learning}
F{\"o}ldi{\'a}k, P. (1991).
\newblock Learning invariance from transformation sequences.
\newblock {\em Neural Computation}, 3(2):194--200.

\bibitem[\protect\astroncite{Franzius et~al.}{2007}]{franzius2007slowness}
Franzius, M., Sprekeler, H., and Wiskott, L. (2007).
\newblock Slowness and sparseness lead to place, head-direction, and
  spatial-view cells.
\newblock {\em PLoS Computational Biology}, 3(8):e166.

\bibitem[\protect\astroncite{Gisslen et~al.}{2011}]{BigDog2011agi}
Gisslen, L., Luciw, M., Graziano, V., and Schmidhuber, J. (2011).
\newblock Sequential constant size compressors for reinforcement learning.
\newblock In {\em {Fourth Conference on Artificial General Intelligence
  (AGI)}}.

\bibitem[\protect\astroncite{Grossberg}{1980}]{grossberg1980does}
Grossberg, S. (1980).
\newblock How does a brain build a cognitive code?.
\newblock {\em Psychological Review}, 87(1):1.

\bibitem[\protect\astroncite{Hafting et~al.}{2005}]{hafting2005microstructure}
Hafting, T., Fyhn, M., Molden, S., Moser, M., and Moser, E. (2005).
\newblock Microstructure of a spatial map in the entorhinal cortex.
\newblock {\em Nature}, 7052:801.

\bibitem[\protect\astroncite{Hinton}{1989}]{Hinton1989}
Hinton, G. (1989).
\newblock Connectionist learning procedures.
\newblock {\em Artificial Intelligence}, 40(1-3):185--234.

\bibitem[\protect\astroncite{Hinton}{2002}]{hinton:2002}
Hinton, G.~E. (2002).
\newblock Training products of experts by minimizing contrastive divergence.
\newblock {\em Neural Comp.}, 14(8):1771--1800.

\bibitem[\protect\astroncite{Jenkins and Matari{\'c}}{2004}]{jenkins2004spatio}
Jenkins, O. and Matari{\'c}, M. (2004).
\newblock A spatio-temporal extension to isomap nonlinear dimension reduction.
\newblock In {\em Proceedings of the twenty-first international conference on
  Machine learning}, page~56. ACM.

\bibitem[\protect\astroncite{Jolliffe}{1986}]{Jolliffe}
Jolliffe, I.~T. (1986).
\newblock {\em Principal Component Analysis}.
\newblock Springer-Verlag, New York.

\bibitem[\protect\astroncite{Klapper-Rybicka et~al.}{2001}]{Klapper:01}
Klapper-Rybicka, M., Schraudolph, N.~N., and Schmidhuber, J. (2001).
\newblock Unsupervised learning in {LSTM} recurrent neural networks.
\newblock In {\em Lecture Notes on Comp. Sci. 2130, Proc. Intl. Conf. on
  Artificial Neural Networks (ICANN-2001)}, pages 684--691. Springer: Berlin,
  Heidelberg.

\bibitem[\protect\astroncite{Kohonen}{2001}]{Kohonen01}
Kohonen, T. (2001).
\newblock {\em Self-Organizing Maps}.
\newblock Springer-Verlag, Berlin, 3rd edition.

\bibitem[\protect\astroncite{Kompella et~al.}{2011a}]{kompellaincremental}
Kompella, V., Luciw, M., and Schmidhuber, J. (2011a).
\newblock Incremental slow feature analysis.
\newblock In {\em International Joint Conference of Artificial Intelligence}.

\bibitem[\protect\astroncite{Kompella et~al.}{2011b}]{AutoIncSFA2011}
Kompella, V.~R., Pape, L., Masci, J., Frank, M., and Schmidhuber, J. (2011b).
\newblock Autoincsfa and vision-based developmental learning for humanoid
  robots.
\newblock In {\em IEEE-RAS International Conference on Humanoid Robots}, Bled,
  Slovenia.

\bibitem[\protect\astroncite{Krasulina}{1970}]{krasulina1970method}
Krasulina, T. (1970).
\newblock Method of stochastic approximation in the determination of the
  largest eigenvalue of the mathematical expectation of random matrices.
\newblock {\em Automat. Remote Contr}, 2:215--221.

\bibitem[\protect\astroncite{Kreyszig}{1988}]{Kreyszig88}
Kreyszig, E. (1988).
\newblock {\em Advanced engineering mathematics}.
\newblock Wiley, New York.

\bibitem[\protect\astroncite{Lee and Seung}{1999}]{lee1999learning}
Lee, D. and Seung, H. (1999).
\newblock Learning the parts of objects by non-negative matrix factorization.
\newblock {\em Nature}, 401(6755):788--791.

\bibitem[\protect\astroncite{Lee et~al.}{2010}]{lee2010unsupervised}
Lee, H., Largman, Y., Pham, P., and Ng, A. (2010).
\newblock Unsupervised feature learning for audio classification using
  convolutional deep belief networks.
\newblock {\em Advances in neural information processing systems},
  22:1096--1104.

\bibitem[\protect\astroncite{Legenstein
  et~al.}{2010}]{legenstein2010reinforcement}
Legenstein, R., Wilbert, N., and Wiskott, L. (2010).
\newblock Reinforcement learning on slow features of high-dimensional input
  streams.
\newblock {\em PLoS Computational Biology}, 6(8).

\bibitem[\protect\astroncite{Lindst\"{a}dt}{1993}]{Steffi:93cmss}
Lindst\"{a}dt, S. (1993).
\newblock Comparison of two unsupervised neural network models for redundancy
  reduction.
\newblock In Mozer, M.~C., Smolensky, P., Touretzky, D.~S., Elman, J.~L., and
  Weigend, A.~S., editors, {\em Proc. of the 1993 Connectionist Models Summer
  School}, pages 308--315. Hillsdale, NJ: Erlbaum Associates.

\bibitem[\protect\astroncite{Mitchison}{1991}]{mitchison1991removing}
Mitchison, G. (1991).
\newblock Removing time variation with the anti-hebbian differential synapse.
\newblock {\em Neural Computation}, 3(3):312--320.

\bibitem[\protect\astroncite{Oja}{1982}]{oja1982simplified}
Oja, E. (1982).
\newblock Simplified neuron model as a principal component analyzer.
\newblock {\em Journal of mathematical biology}, 15(3):267--273.

\bibitem[\protect\astroncite{Oja}{1985}]{Oja85}
Oja, E. (1985).
\newblock On stochastic approximation of the eigenvectors and eigenvalues of
  the expectation of a random matrix.
\newblock {\em Journal of Mathematical Analysis and Applications}, 106:69--84.

\bibitem[\protect\astroncite{Oja}{1992}]{oja1992principal}
Oja, E. (1992).
\newblock Principal components, minor components, and linear neural networks.
\newblock {\em Neural Networks}, 5(6):927--935.

\bibitem[\protect\astroncite{O'Keefe and Dostrovsky}{1971}]{o1971hippocampus}
O'Keefe, J. and Dostrovsky, J. (1971).
\newblock The hippocampus as a spatial map: Preliminary evidence from unit
  activity in the freely-moving rat.
\newblock {\em Brain research}.

\bibitem[\protect\astroncite{Papoulis et~al.}{1965}]{papoulis1965probability}
Papoulis, A., Pillai, S., and Unnikrishna, S. (1965).
\newblock {\em Probability, random variables, and stochastic processes}, volume
  196.
\newblock McGraw-hill New York.

\bibitem[\protect\astroncite{Peng and Yi}{2006}]{peng2006new}
Peng, D. and Yi, Z. (2006).
\newblock A new algorithm for sequential minor component analysis.
\newblock {\em International Journal of Computational Intelligence Research},
  2(2):207--215.

\bibitem[\protect\astroncite{Peng et~al.}{2007}]{peng2007convergence}
Peng, D., Yi, Z., and Luo, W. (2007).
\newblock Convergence analysis of a simple minor component analysis algorithm.
\newblock {\em Neural Networks}, 20(7):842--850.

\bibitem[\protect\astroncite{Rolls}{1999}]{rolls1999spatial}
Rolls, E. (1999).
\newblock Spatial view cells and the representation of place in the primate
  hippocampus.
\newblock {\em Hippocampus}, 9(4):467--480.

\bibitem[\protect\astroncite{Sanger}{1989}]{sanger1989optimal}
Sanger, T. (1989).
\newblock Optimal unsupervised learning in a single-layer linear feedforward
  neural network.
\newblock {\em Neural networks}, 2(6):459--473.

\bibitem[\protect\astroncite{Schmidhuber}{1992a}]{Schmidhuber:92ncchunker}
Schmidhuber, J. (1992a).
\newblock Learning complex, extended sequences using the principle of history
  compression.
\newblock {\em Neural Computation}, 4(2):234--242.

\bibitem[\protect\astroncite{Schmidhuber}{1992b}]{Schmidhuber:92ncfactorial}
Schmidhuber, J. (1992b).
\newblock Learning factorial codes by predictability minimization.
\newblock {\em Neural Computation}, 4(6):863--879.

\bibitem[\protect\astroncite{Schmidhuber}{1992c}]{Schmidhuber:92nips}
Schmidhuber, J. (1992c).
\newblock Learning unambiguous reduced sequence descriptions.
\newblock In Moody, J.~E., Hanson, S.~J., and Lippman, R.~P., editors, {\em
  Advances in Neural Information Processing Systems 4 (NIPS 4)}, pages
  291--298. Morgan Kaufmann.

\bibitem[\protect\astroncite{Schmidhuber}{1999}]{Schmidhuber:99zif}
Schmidhuber, J. (1999).
\newblock Neural predictors for detecting and removing redundant information.
\newblock In Cruse, H., Dean, J., and Ritter, H., editors, {\em Adaptive
  Behavior and Learning}. Kluwer.

\bibitem[\protect\astroncite{Schmidhuber}{2010}]{schmidhuber2010formal}
Schmidhuber, J. (2010).
\newblock Formal theory of creativity, fun, and intrinsic motivation
  (1990--2010).
\newblock {\em IEEE Transactions on Autonomous Mental Development},
  2(3):230--247.

\bibitem[\protect\astroncite{Sprekeler et~al.}{2010}]{sprekeler2010extension}
Sprekeler, H., Zito, T., and Wiskott, L. (2010).
\newblock An extension of slow feature analysis for nonlinear blind source
  separation.

\bibitem[\protect\astroncite{T.~Zito and Berkes}{2008}]{MDP2008}
T.~Zito, N.~Wilbert, L.~W. and Berkes, P. (2008).
\newblock Modular toolkit for data processing (mdp): a python data processing
  framework.
\newblock {\em Frontiers in Neuroinformatics}, 2.

\bibitem[\protect\astroncite{Taube et~al.}{1990}]{taube1990head}
Taube, J., Muller, R., and Ranck, J. (1990).
\newblock Head-direction cells recorded from the postsubiculum in freely moving
  rats. i. description and quantitative analysis.
\newblock {\em The Journal of Neuroscience}, 10(2):420.

\bibitem[\protect\astroncite{V.~Tikhanoff and Nori}{2008}]{iCubSim}
V.~Tikhanoff, A.~Cangelosi, P. F. G. M. L.~N. and Nori, F. (2008).
\newblock An open-source simulator for cognitive robotics research: The
  prototype of the icub humanoid robot simulator.

\bibitem[\protect\astroncite{Wang and Karhunen}{1996}]{wang1996unified}
Wang, L. and Karhunen, J. (1996).
\newblock A unified neural bigradient algorithm for robust pca and mca.
\newblock {\em International journal of neural systems}, 7(1):53.

\bibitem[\protect\astroncite{Weng and Zhang}{2006}]{weng2006optimal}
Weng, J. and Zhang, N. (2006).
\newblock Optimal in-place learning and the lobe component analysis.
\newblock In {\em Neural Networks, 2006. IJCNN'06. International Joint
  Conference on}, pages 3887--3894. IEEE.

\bibitem[\protect\astroncite{Weng et~al.}{2003}]{WengCCIPCA03}
Weng, J., Zhang, Y., and Hwang, W. (2003).
\newblock Candid covariance-free incremental principal component analysis.
\newblock 25(8):1034--1040.

\bibitem[\protect\astroncite{Wiskott}{2003}]{wiskott2003estimating}
Wiskott, L. (2003).
\newblock Estimating driving forces of nonstationary time series with slow
  feature analysis.
\newblock {\em Arxiv preprint cond-mat/0312317}.

\bibitem[\protect\astroncite{Wiskott et~al.}{2011}]{sfascholar}
Wiskott, L., Berkes, P., Franzius, M., Sprekeler, H., and Wilbert, N. (2011).
\newblock Slow feature analysis.
\newblock {\em Scholarpedia}, 6(4):5282.

\bibitem[\protect\astroncite{Wiskott and Sejnowski}{2002}]{WisSej2002}
Wiskott, L. and Sejnowski, T. (2002).
\newblock Slow feature analysis: Unsupervised learning of invariances.
\newblock {\em Neural Computation}, 14(4):715--770.

\bibitem[\protect\astroncite{Xu et~al.}{1992}]{xu1992modified}
Xu, L., Oja, E., and Suen, C. (1992).
\newblock Modified hebbian learning for curve and surface fitting.
\newblock {\em Neural Networks}, 5(3):441--457.

\bibitem[\protect\astroncite{Zhang and Weng}{2001}]{zhang2001convergence}
Zhang, Y. and Weng, J. (2001).
\newblock Convergence analysis of complementary candid incremental principal
  component analysis.
\newblock {\em Michigan State University}.

\end{thebibliography}
\end{document}